\documentclass{article}
\usepackage{spconf,amsmath,graphicx}

\usepackage{url}
\usepackage{bbding, makecell}
\usepackage{hyperref}
\usepackage{amsfonts}
\usepackage{romannum}
\usepackage{makecell}
\usepackage{multirow}
\usepackage{xcolor}  
\usepackage{colortbl}  
\usepackage{threeparttable}
\usepackage{pgfplots}  
 
\newcommand{\etal}{\textit{et al.}}

\title{SSDA-YOLO: Semi-supervised Domain Adaptive YOLO for Cross-Domain Object Detection}

\name{Huayi Zhou$^{\star}$ \qquad Fei Jiang$^{\dagger}$ \qquad Hongtao Lu$^{\star}$}
\address{$^{\star}$ Shanghai Jiao Tong University, sjtu\_zhy@sjtu.edu.cn, htlu@sjtu.edu.cn; \\
    $^{\dagger}$ East China Normal University, fjiang@mail.ecnu.edu.cn}

\begin{document}
%
\maketitle
%
\begin{abstract}
Domain adaptive object detection (DAOD) aims to alleviate transfer performance degradation caused by the cross-domain discrepancy. However, most existing DAOD methods are dominated by outdated and computationally intensive two-stage Faster R-CNN, which is not the first choice for industrial applications. In this paper, we propose a novel semi-supervised domain adaptive YOLO (SSDA-YOLO) based method to improve cross-domain detection performance by integrating the compact one-stage stronger detector YOLOv5 with domain adaptation. Specifically, we adapt the knowledge distillation framework with the Mean Teacher model to assist the student model in obtaining instance-level features of the unlabeled target domain. We also utilize the scene style transfer to cross-generate pseudo images in different domains for remedying image-level differences. In addition, an intuitive consistency loss is proposed to further align cross-domain predictions. We evaluate SSDA-YOLO on public benchmarks including PascalVOC, Clipart1k, Cityscapes, and Foggy Cityscapes. Moreover, to verify its generalization, we conduct experiments on yawning detection datasets collected from various real classrooms. The results show considerable improvements of our method in these DAOD tasks, which reveals both the effectiveness of proposed adaptive modules and the urgency of applying more advanced detectors in DAOD. Our code is available on \url{https://github.com/hnuzhy/SSDA-YOLO}.
\end{abstract}
%

\section{Introduction}

Modern object detection methods based on convolution neural networks (CNNs) have achieved many remarkable improvements~\cite{ren2015faster, tian2019fcos, liu2016ssd, he2017mask, lin2017feature, bochkovskiy2020yolov4, ge2021yolox}. However, the high accuracy of these methods are mostly restricted to the training set source domain. Thus, even if state-of-the-art methods achieve superb results on large scale benchmarks (e.g., PascalVOC~\cite{everingham2010pascal}, MSCOCO~\cite{lin2014microsoft}, and OpenImages~\cite{kuznetsova2020open}), significant performance degradation often arises when testing under the very different target domain scenario. The new target scene may include dissimilar image styles, lighting conditions, image quality, camera perspectives, etc., which often bring considerable domain shifts between the training and test data. Although collecting more training data can alleviate this problem, it is impractical for the expensive and time-consuming labeling process. In some scenarios, e.g. biomedical image, it is even impossible to get extensive precise annotations.

Addressing the domain gap between source training datasets and target testing datasets is the focus of domain adaptive object detection (DAOD)~\cite{chen2018domain, saito2019strong, deng2021unbiased, khodabandeh2019robust, yao2021multi, li2022sigma}. Generally, the DAOD attempts to learn a robust and generalizable detector using labeled data from the source domain and unlabeled data from the target domain. The domain adaptive Faster R-CNN~\cite{chen2018domain} is a milestone study developed for tackling the domain shift problem in object detection. Following \cite{chen2018domain}, most domain adaptation approaches~\cite{saito2019strong, deng2021unbiased, khodabandeh2019robust, yao2021multi, li2022sigma} are still based on the Faster R-CNN. Instead of using the two-stage detector Faster R-CNN with ROIs for more convenient local adaptation, some recent works~\cite{hsu2020every, chen2021i3net, li2022sigma, zhou2022multi} have proposed one-stage detector based DAOD methods considering its computational advantage. Following YOLO series~\cite{redmon2018yolov3, bochkovskiy2020yolov4, jocher2020yolov5}, some more light-weight DAOD methods are proposed~\cite{zhang2021domain, liu2021image, hnewa2021multiscale, vidit2021attention}. Generally, the particular domain adaptive solution is inseparable from the breakthrough in object detection methods and architectures.

Towards resource-limited and time-critical real applications, we adopt the current widely used YOLOv5~\cite{jocher2020yolov5} as the basic object detector in our framework. There are two actualities motivate us. On one hand, one-stage object detectors, YOLO series particularly (e.g., YOLOv5~\cite{jocher2020yolov5} and YOLOX~\cite{ge2021yolox}), can reach almost real-time with maintaining comparable or superior accuracy as two-stage detectors. This makes them invaluable for time-sensitive scenes like autonomous driving and action recognition. On the other hand, outdated Faster R-CNN based methods have dominated the DAOD field. We doubt whether it will limit or mislead the overall performance of DAOD algorithms. Although the advanced stronger detector YOLOv5 has shown a compelling balance of high performance and time-consuming currently, few researches explore the introduction of the YOLO architecture. Thus, we expect to excavate and migrate the superiority of YOLOv5 in DAOD.

\begin{figure}[]
	\centering
	\includegraphics[width=1\columnwidth]{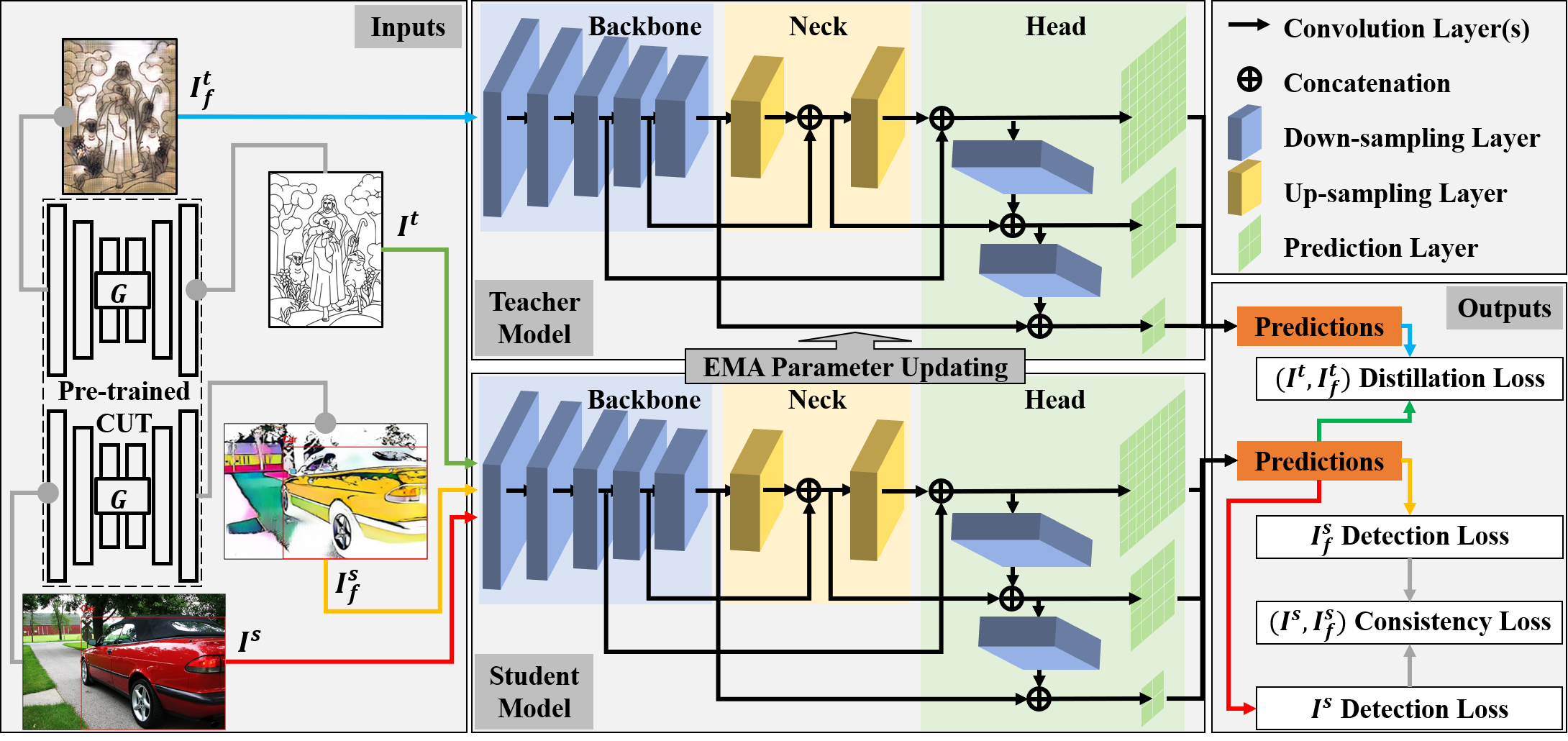}
	\caption{The overall architecture of our proposed SSDA-YOLO. {\bf Middle:} We adopt the original YOLOv5~\cite{jocher2020yolov5} network as the basic detector for both teacher and student models in a knowledge distillation framework. {\bf Left:} Besides taking the real source image $\mathbf{I}^s$ and target image $\mathbf{I}^t$ as inputs during training, we have also generated the target-like fake source image $\mathbf{I}^s_f$ and the source-like fake target image $\mathbf{I}^t_f$ using corresponding pre-trained CUT~\cite{park2020contrastive} models to alleviate image-level domain differences. {\bf Right:} Based on predictions of multiple inputs (with different colors), we have constructed various corresponding loss functions for supporting the semi-supervised learning.}
	\label{fig1}
\end{figure}

In this paper, we propose a novel semi-supervised domain adaptive YOLO (SSDA-YOLO) method. With the YOLOv5 as the backbone network, SSDA-YOLO can extract source domain features efficiently by fully supervised learning. In order to obtain {\it instance-level} features of the target domain, we adopt the knowledge distillation framework, and use the Mean Teacher~\cite{tarvainen2017mean} guided teacher network to detect unlabeled target images. Then, we filter the predictions to generate strong pseudo-labels iteratively for enforcing a relatively unbiased updating of the student network. Furthermore, to narrow the distance between the teacher and student model in {\it image-level}, we use the superior unpaired image generation approach CUT~\cite{park2020contrastive} to synthesize pseudo images offline as additional inputs. The overall architecture is shown in Fig.~\ref{fig1}. To verify the effectiveness of SSDA-YOLO, we have conducted extensive experiments using both public benchmarks and yawning detection datasets extracted from real classroom scenarios. Experiment results show significant improvements on each target domain test set.

In summary, our contributions are as follows: 1) We propose a novel semi-supervised domain adaptive YOLO (SSDA-YOLO) for tackling the DAOD problem, which combines the stronger one-stage detector YOLOv5 with a knowledge distillation framework. 2) Two newly designed domain adaptive penalty functions including the distillation loss and the consistency loss are proven reasonable and effective. 3) Our method can achieve comparable or superior accuracy improvements on two popular domain transfer experiments (e.g., {\bf PascalVOC$\mapsto$Clipart1k}, and {\bf Cityscapes$\mapsto$Foggy Cityscapes}), and maintain appreciable generalization in yawning detection under different classroom scenarios. 4) We expose and validate the necessity and urgency of using more advanced detectors in the DAOD field, which is trapped by the outdated detector Faster R-CNN.

\section{Related Works}

\subsection{Object Detection}

Modern object detection methods can be grouped into two categories: 1) The {\it two-stage} architectures (e.g., Faster R-CNN~\cite{ren2015faster} and Mask R-CNN~\cite{he2017mask}), that first extract regions of interest (ROIs) and then operate bounding boxes classification and regression; and 2) {\it one-stage} detectors (e.g., SSD~\cite{liu2016ssd}, FPN~\cite{lin2017feature}, FCOS~\cite{tian2019fcos}, and YOLO series \cite{redmon2018yolov3, bochkovskiy2020yolov4, jocher2020yolov5, ge2021yolox}), that directly output bounding boxes and classes from the predicted feature maps with pre-defined anchors. While the former yields slightly higher accuracy, the latter is faster and more compact, which makes them better suited for time-sensitive applications and computation limited edge devices. Recently, the success of Transformers ~\cite{dosovitskiy2020image} applied for image recognition, which has been shown to be more robust than the predominant CNNs~\cite{bhojanapalli2021understanding}, is inherited to design various end-to-end DEtection TRansformers (DETR)~\cite{carion2020end, zhu2020deformable, liu2022dab, li2022dn}. Despite achieving high detection accuracy, DETRs still suffer from sophisticated architectures and slow convergence issues.

In particular, we choose the simple yet efficient YOLOv5, which combines CSPNet~\cite{wang2020cspnet}, FPN~\cite{lin2017feature} and Focal loss~\cite{lin2017focal}, as our basic detector. Benefits from the Cross-Stage-Partial (CSP) bottlenecks~\cite{wang2020cspnet} and optimizations in training tricks, the one-stage YOLO series even surpass their two-stage counterparts both in accuracy and efficiency. Beyond that, it is proven having effective data augmentations (e.g., Mosaic and MixUp) and strong versatility during training.

\subsection{Cross-Domain Object Detection}

Existing cross-domain object detection methods have exclusively focused on the two-stage detector Faster R-CNN~\cite{chen2018domain, saito2019strong, he2019multi, shen2019scl, khodabandeh2019robust, kim2019diversify, he2020domain, nguyen2020domain, he2021partial, xu2020exploring, chen2020harmonizing, deng2021unbiased, yao2021multi, shi2022universal, li2022sigma, zhao2022task, wu2022target, zhou2022multi, he2022cross, chen2022learning}. Following the pioneering DA-Faster~\cite{chen2018domain} which introduces the Gradient Reversal Layer (GRL)~\cite{ganin2016domain} and firstly designs instance-level and image-level alignments to promote performance on new domains, SWDA~\cite{saito2019strong} proposes similar strong local and weak global feature alignments for improvement. Then, SCL~\cite{shen2019scl} presents a gradient detach based stacked complementary losses method which also uses the Faster R-CNN as detector. NLDA~\cite{khodabandeh2019robust} formulates a robust learning as training with noisy labels on the target domain. Meanwhile, He \etal~have designed multi-adversarial Faster R-CNN (MAF~\cite{he2019multi}), asymmetric tri-way Faster R-CNN (ATF~\cite{he2020domain}) and partial alignment based ATF (PA-ATF~\cite{he2021partial}) one after another. MEAA~\cite{nguyen2020domain} proposes a Faster R-CNN based method consisting of two tailor-made modules named local uncertainty attentional alignment (LUAA) and multi-level uncertainty-aware context alignment (MUCA). UMT~\cite{deng2021unbiased} generates fake training images using CycleGAN~\cite{zhu2017unpaired} to alleviate domain bias. Focusing on labeled data from multiple source domains, MSDA~\cite{yao2021multi} proposes the Divide-and-Merge Spindle Network (DMSN) to enhance domain invariance and preserve discriminative power. US-DAF~\cite{shi2022universal} implements a universal scale-aware domain adaptive Faster R-CNN with multi-label learning for reducing the negative transfer effect during training. SIGMA~\cite{li2022sigma} represents the source and target data as graphs, and reformulates the adaptation as a graph matching problem.

In comparison, some recent works try to tackle the DAOD problem using one-stage detectors~\cite{hsu2020every, chen2021i3net, liu2021image, hnewa2021multiscale, vidit2021attention}. For example, EPMDA~\cite{hsu2020every} adapts FCOS~\cite{tian2019fcos} to explicitly extracts objectness maps. I$^3$Net~\cite{chen2021i3net} introduces complementary modules specifically designed for the SSD~\cite{liu2016ssd} architecture. And~\cite{zhang2021domain, hnewa2021multiscale, vidit2021attention} have designed pertinent methods and frameworks based on YOLOv3/4/5 respectively.

In general, DAOD is chronically dominated by the outdated two-stage Faster R-CNN, which contains region proposals for most methods to effortlessly perform local adaptation. This has been shown to significantly improve the adaptation effectiveness. We know that they follow Faster R-CNN for easy comparison with previously published DAOD works. However, even their strongest architecture (Faster R-CNN+ResNet101), comparing to recently proposed simple yet more efficient YOLOv5 and YOLOX, is far less efficient and accurate in the area of supervised object detection. A natural question is whether outdated detectors will limit or mislead the overall performance of DAOD algorithms. Our hypothesis is that both the basic detectors and appropriate training strategies play important roles for DAOD. Actually, the recently proposed DAFormer~\cite{hoyer2022daformer} for tackling UDA semantic segmentation task also pointed out the similar problem, that DA semantic segmentation is dominated by outdated segmentators DeepLabV2~\cite{chen2017deeplab} and FCN8s~\cite{long2015fully}. It used sophisticated and more robust Transformers~\cite{dosovitskiy2020image} and SegFormer~\cite{xie2021segformer} instead of the predominant CNNs to enhance the UDA performance. Their experiments confirmed the advantages of superseding DeepLabV2 with Transformers. Similarly, we explore to break the unreasonable status quo that Faster R-CNN dominates DAOD researches, propose our method SSDA-YOLO combining the single-stage detector YOLOv5 and several corresponding adaptive modules, and reveal the same discipline that the architecture choice can limit the effectiveness of UDA methods.

\subsection{Semi-supervised Domain Adaptation}

Unsupervised domain adaptation (UDA) is defined to adapt a model from the labeled source domain to an unlabeled target domain. It is first widely studied for image classification tasks~\cite{duan2012domain, gopalan2011domain, long2015learning, ganin2015unsupervised, motiian2017unified}. For the DAOD problem in this paper, by default, labels of the target domain are invisible during training, but only images are used. Generally, in order to learn domain-invariant representations and minimize the distance metric among domains, almost all previous DAOD methods~\cite{chen2018domain, saito2019strong, hsu2020every, chen2021i3net, li2022sigma, zhao2022task, zhou2022multi} deal with images in the source and target domains separately.

Moreover, in practical terms, we can effortlessly get a small partial labeled images in the target scenario. Then, we can obtain considerable benefits by applying the semi-supervised learning (or called few-shot learning). Therefore, beyond the general UDA setting, DTPL~\cite{inoue2018cross} proposes a {\it weakly supervised} progressive domain adaptation framework by providing image-level annotations of target domain images. MTOR~\cite{cai2019exploring} firstly explores object relation in Mean Teacher (MT)~\cite{tarvainen2017mean} which is initially designed for {\it semi-supervised} learning task. Then, UMT~\cite{deng2021unbiased} promotes the Faster R-CNN adaptation by designing and utilizing an unbiased Mean Teacher. Recently, TRKP~\cite{wu2022target}, TDD~\cite{he2022cross} and PT~\cite{chen2022learning} apply the knowledge distillation framework with applying the MT model to remedy cross-domain discrepancies and perceive target-relevant features. DAFormer~\cite{hoyer2022daformer} for tackling the cross-domain semantic segmentation task also adopted the MT model in their self-training pipeline.

Inspired by these principles, we construct our method in a knowledge distillation structure with integrating the prevalent MT model as shown in Fig.~\ref{fig1}. It utilizes supervised learning in the source dataset, and performs unsupervised learning in the target dataset. Besides, unlabeled target training images are style-translated with source-like global scene before fed in teacher model. This combination solution forms our semi-supervised domain adaptation SSDA-YOLO.


\section{Preliminaries and Motivation}

As discussed above, most state-of-the-art DAOD methods are confined to two-stage detectors, especially the outdated Faster R-CNN. This is largely because the Faster R-CNN provides two clear branches of classification and localization. DA-Faster~\cite{chen2018domain} takes advantage of this characteristic, and firstly proposes two representations at instance-level and image-level which are widely followed until now. Recently, one-stage detectors based DAOD methods are also considering how to extract cross-domain features with their designed adaptation components at these two levels. Besides, most DAOD methods focus on the training of one shared detection network with data from both source and target domains. This adversarial way is distressful to optimize and converge. These premises motivate us in addressing two principal challenges:


{\bf Knowledge Distillation Structure:} The previous DAOD method using a single shared network to fit the cross-domain data is an adversarial process. Most of them utilize the bidirectional operator Gradient Reversal Layer (GRL)~\cite{ganin2016domain} which is used to realize two conflicted optimization objectives. On one hand, it acts as an identity operator for minimizing the classification error during forward training. On the other hand, it becomes a negative scalar during back-propagation for maximizing the binary-classification error and learning domain-invariant features. Distance metrics like Maximum Mean Discrepancy (MMD)~\cite{gretton2012kernel} is usually applied to measure the domain shift and supervise the model. Despite these common settings, the more robust teacher-student framework is adopted in DAOD by recently proposed approaches~\cite{deng2021unbiased, wu2022target, he2022cross, chen2022learning}. Their distillation structures can enhance source detector to perceive objects in a target image. However, these methods are all based on the Faster R-CNN, and distill intermediate features based knowledge~\cite{gou2021knowledge}. We also follow the teacher-student framework, but distill knowledge based on final response of superior one-stage detector YOLOv5. Among various model setups of teacher and student relationship~\cite{gou2021knowledge}, we choose to maintain them the identical architecture.


{\bf Cross-domain Features Extraction:} In the one-stage detection framework, it has unified predictions and indistinct discrimination between classification and localization. For example, EPMDA~\cite{hsu2020every} based on FCOS~\cite{tian2019fcos} proposes the global and center-aware discriminators to imitate image-level and instance-level features extraction. I$^3$Net~\cite{chen2021i3net} based on SSD~\cite{liu2016ssd} designs a multi-label classifier and two domain discriminators to compensate image-level and pixel-level features. The most similar to our work are DA-YOLO~\cite{zhang2021domain} and MS-DAYOLO~\cite{hnewa2021multiscale}. DA-YOLO based on YOLOv3~\cite{redmon2018yolov3} proposes two adaptive modules RIA and MSIA using three domain classifiers to perform image and instance level adaptation respectively. MS-DAYOLO based on YOLOv4~\cite{bochkovskiy2020yolov4} attaches a domain adaptation network (DAN) to the backbone to directly learn domain-invariant features in multiscale. Unlike all of them, we adopt the pseudo cross-generated images to address image-level shifts, and the Mean Teacher model to obtain target domain features at instance-level for guiding the student model training.


\section{Proposed Method}

Our proposed SSDA-YOLO method is based on the advanced one-stage detector YOLOv5. It contains four main components: the Mean Teacher model with a knowledge distillation framework for guiding robust student network updating, the pseudo cross-generated training images for alleviating image-level domain differences, the updated distillation loss for remedying cross-domain discrepancy, and the novel consistency loss for further redressing cross-domain objectness bias. Details of these modules are as follows.

\subsection{Definition of Terms}

For the cross-domain object detection task, we have a set of source images $\mathbf{I}^s$ annotated with totally $N$ object bounding boxes $\mathcal{B}=\{{\mathit{B}_j\mid}^N_{j=1}, \mathit{B}_j=(x_j,y_j,w_j,h_j) \}$ and corresponding class labels $\mathcal{C}=\{ {\mathit{C}_j\mid}^N_{j=1}, \mathit{C}_j\in(0,1,...,c) \}$ with $c$ object classes, and a set of unlabeled target images $\mathbf{I}^t$. With $\mathit{N_s}$ source images $\mathbf{I}^s$, also the bounding box coordinates set $\mathcal{B}^s$ and the class labels set $\mathcal{C}^s$, we represent the source domain as $\mathcal{D}_s=\{ (\mathbf{I}^s_i, \mathcal{B}^s_i, \mathcal{C}^s_i)\mid^{\mathit{N}_s}_{i=1} \}$. Similarly, we define the target domain with $\mathit{N_t}$ label invisible images as $\mathcal{D}_t=\{ \mathbf{I}^t_i \mid ^{\mathit{N}_t}_{i=1}\}$.

Our goal of solving the DAOD problem is to learn a model that could achieve best possible performance for the target domain with given datasets $\mathcal{D}_s$ and  $\mathcal{D}_t$. In particular, we employ YOLOv5 \cite{jocher2020yolov5} as our detector backbone. Following its implements, the loss for training a supervised model with the labeled source dataset $\mathcal{D}_s$ can be written as:
\begin{equation}
  \mathcal{L}_{det}(\mathbf{I}^s, \mathcal{B}^s, \mathcal{C}^s) = \mathcal{L}_{box}(\mathcal{B}^s; \mathbf{I}^s) + \mathcal{L}_{cls, obj}(\mathcal{C}^s; \mathbf{I}^s) ~
  \label{eqn1}
\end{equation}
where $\mathcal{L}_{box}$ is the GIoU loss for predicted bounding boxes, and $\mathcal{L}_{cls, obj}$ is the Focal loss~\cite{lin2017focal} of both classification probability and objectness score calculated by binary cross entropy. We mainly introduce the unsupervised domain adaptation related to the unlabeled target domain images $\mathcal{D}_t$ in below sections.

\subsection{Mean Teacher Model}

The Mean Teacher (MT) model~\cite{tarvainen2017mean} is initially proposed for semi-supervised learning in image classification. It consists of a typical knowledge distillation structure with two identical model architectures (student and teacher). For domain adaptation tasks, the student model is trained with labeled data in source domain using the gradient descent optimizer. According to the MT model setting, the teacher model is updated by the exponential moving average (EMA) weights from the student model. Specifically, supposing that weight parameters of student and teacher models are noted as $\mathcal{P}_s$ and $\mathcal{P}_t$ respectively, we update $\mathcal{P}_t$ at each training batch step as following:
\begin{equation}
	\mathcal{P}_t = \gamma{\mathcal{P}_t} + (1-\gamma){\mathcal{P}_s}~
	\label{eqn2}
\end{equation}
where $\gamma$ is the exponential decay. Its reasonable value is close to 1.0, typically in the multiple-nines range: 0.99, 0.999, etc.

When applying the MT model to our cross-domain object detection task, we set the unlabeled target domain samples $\mathcal{D}_t$ as the single input of the teacher model. We also train the student model partially on these unlabeled samples $\mathbf{I}^t$. During the distillation, by selecting bounding boxes with high probabilities from the teacher model predictions as pseudo labels, the student model tends to reduce variance on the target domain and enhance the model robustness. Supposing that we have the augmented target inputs $\mathbf{\hat{I}}^t$ for the teacher model and $\mathbf{\bar{I}}^t$ for the student model from the same images $\mathbf{I}^t$, the inconsistency in predictions between the two models can be penalized using a distillation loss defined as below:
\begin{equation}
  \mathcal{L}_{dis}(\mathbf{\hat{I}}^t, \mathbf{\bar{I}}^t) = \mathcal{L}_{det}(\mathbf{\bar{I}}^t, \mathcal{G}_\mathcal{B}[\mathcal{F}_\mathcal{B}(\mathbf{\hat{I}}^t)], \mathcal{G}_\mathcal{C}[\mathcal{F}_\mathcal{C}(\mathbf{\hat{I}}^t)]) ~
  \label{eqn3}
\end{equation}
where $\mathcal{F}_\mathcal{B}(\cdot)$ and $\mathcal{F}_\mathcal{C}(\cdot)$ are the prediction branches from the teacher model for bounding box coordinates and object classes with high maximum category score on $\mathbf{\hat{I}}^t$ respectively. The $\mathcal{G}_\mathcal{B}[\cdot]$ and $\mathcal{G}_\mathcal{C}[\cdot]$ are corresponding filters. Specifically, we set the MT model to evaluation mode in each step during training, and use Non-Maximum Suppression (NMS) to filter the predicted bounding boxes sorted by object confidence with a IoU threshold $\tau_{box}$. Then, we select out the bounding boxes with category scores higher than a threshold $\tau_{cls}$. The final pseudo labels provide the student model {\it instance-level}  features of the target domain.

\subsection{Pseudo Training Images Generation}

Although we have successfully constructed a basic distillation network, it is foreseeable that the weights updating of the student model is dominated by images $\mathbf{I}^s$ in the source domain. In contrast, the teacher model does not touch the source images and is guided by the target domain features. We need to alleviate image-level domain differences which cause two models biased toward their monotonous image inputs. Inspired by SWDA~\cite{saito2019strong} which learns domain-invariant features by weak alignment at the global scene-level using CycleGAN~\cite{zhu2017unpaired}, UMT~\cite{deng2021unbiased} also utilizes CycleGAN to transfer source domain images into target-like domain, and vice versa. TDD~\cite{he2022cross} adopts the FDA~\cite{yang2020fda} which is based on the traditional Fourier Transform in its style transfer module to generate target-like images as extra object supervision in the target domain. Following their consensuses, in this paper, we choose to generate both target-like fake source images and source-like fake target images for training.

\begin{figure}[]
\begin{minipage}[b]{.169\linewidth}
  \centering
  \leftline{\includegraphics[width=1\textwidth]{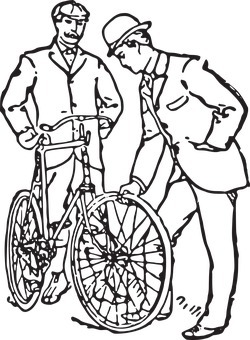}}
  \leftline{\includegraphics[width=1\textwidth]{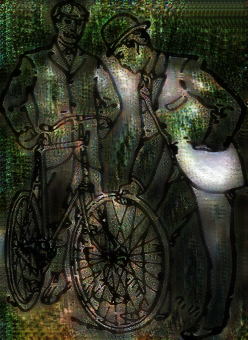}}
  \leftline{\includegraphics[width=1\textwidth]{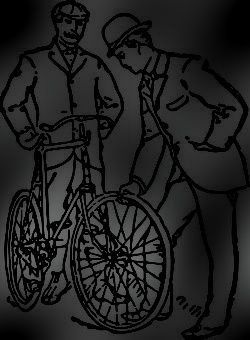}}
  \leftline{\includegraphics[width=1\textwidth]{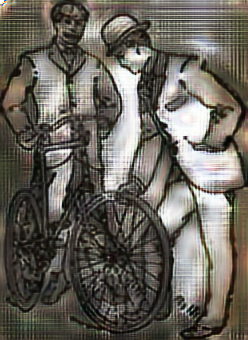}}
  \centerline{\footnotesize{(a)}}\medskip
\end{minipage}
\begin{minipage}[b]{.230\linewidth}
  \centering
  \centerline{\includegraphics[width=1\textwidth]{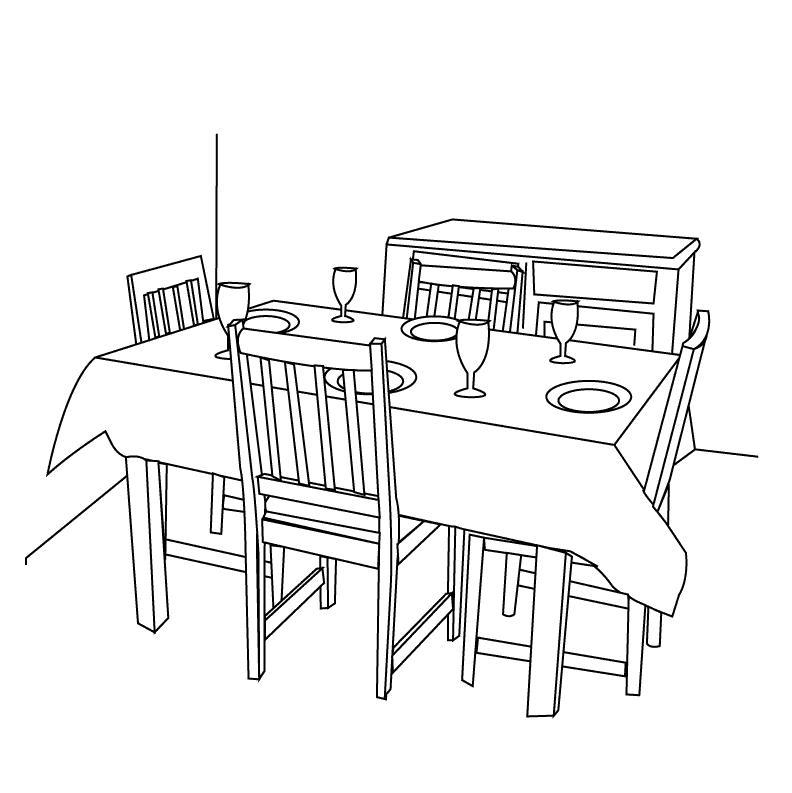}}
  \centerline{\includegraphics[width=1\textwidth]{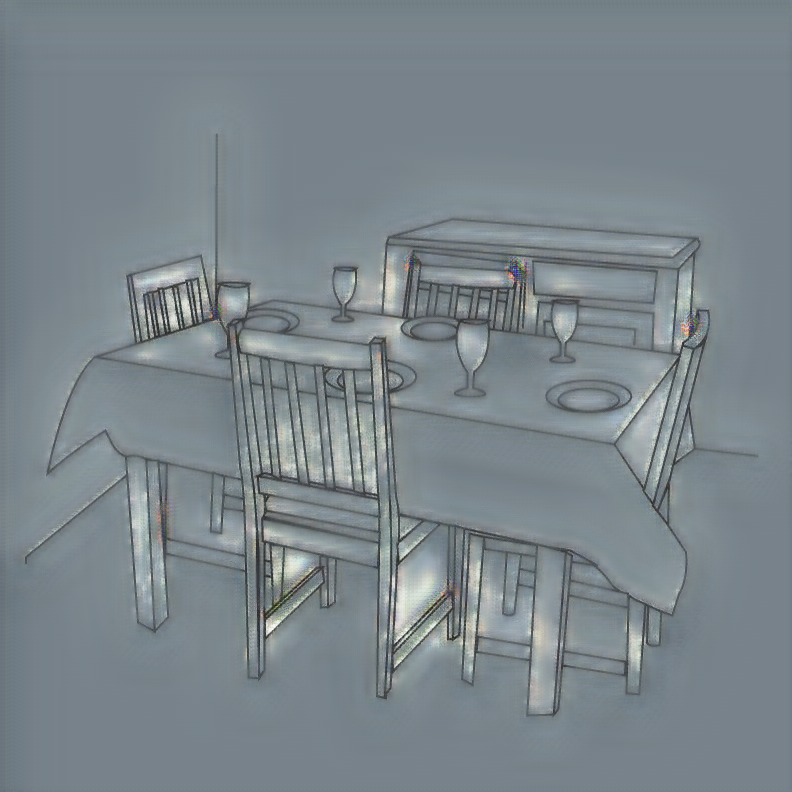}}
  \centerline{\includegraphics[width=1\textwidth]{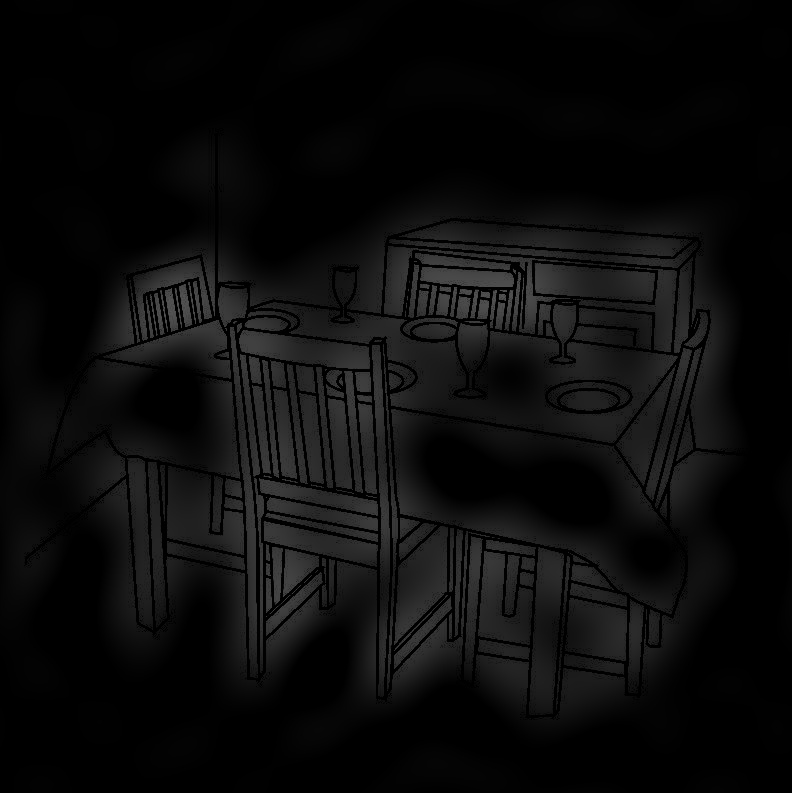}}
  \centerline{\includegraphics[width=1\textwidth]{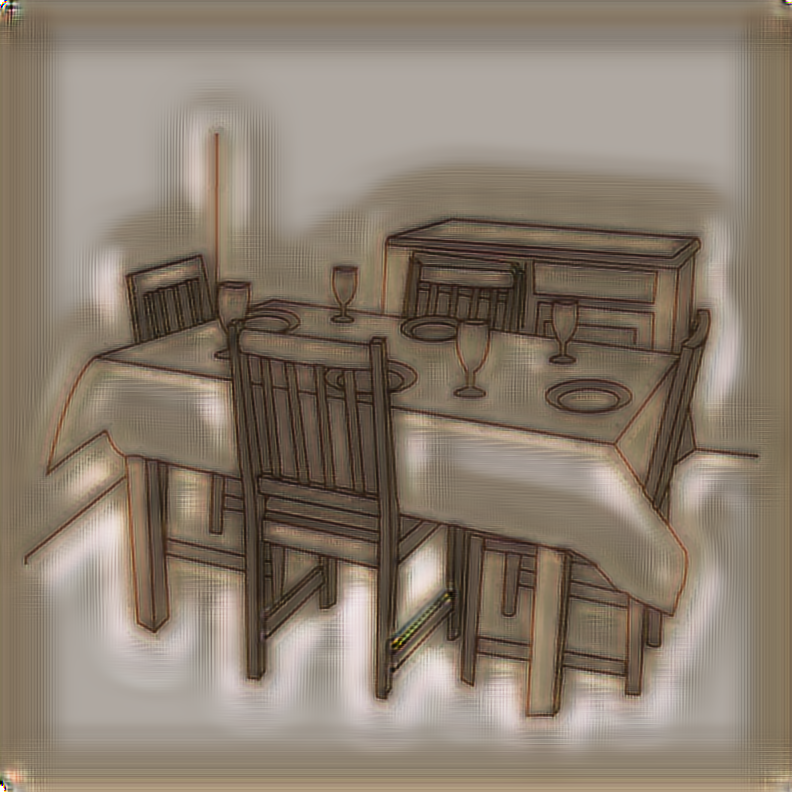}}
  \centerline{\footnotesize{(b)}}\medskip
\end{minipage}
\begin{minipage}[b]{.261\linewidth}
  \centerline{\includegraphics[width=1\textwidth]{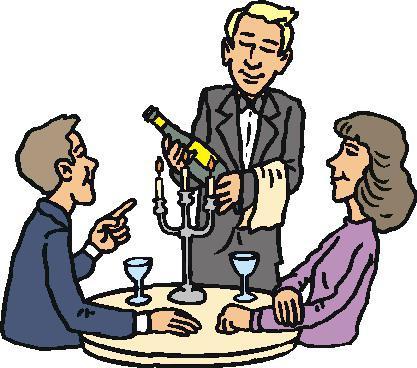}}
  \centerline{\includegraphics[width=1\textwidth]{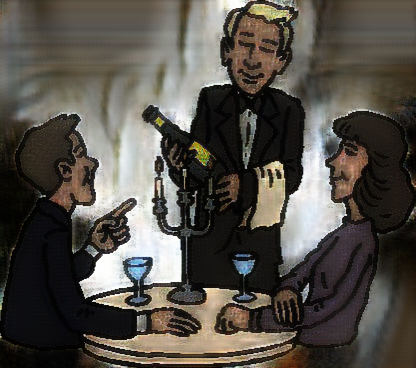}}
  \centerline{\includegraphics[width=1\textwidth]{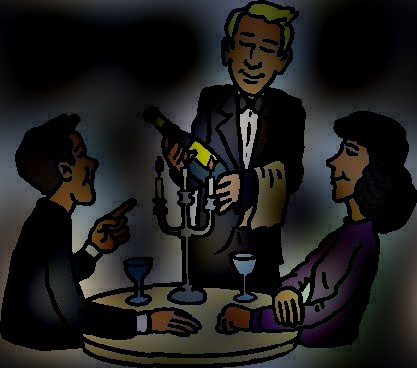}}
  \centerline{\includegraphics[width=1\textwidth]{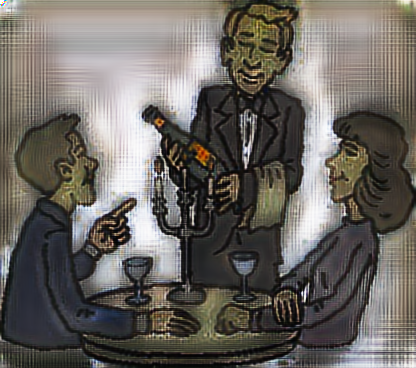}}
  \centerline{\footnotesize{(c)}}\medskip
\end{minipage}
\begin{minipage}[b]{.300\linewidth}
  \rightline{\includegraphics[width=1\textwidth]{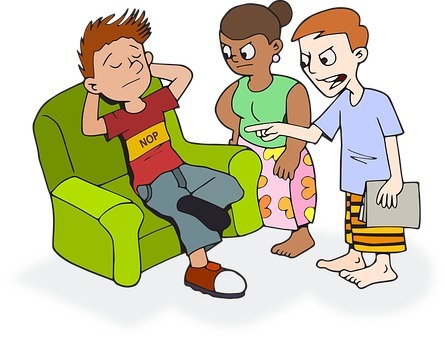}}
  \rightline{\includegraphics[width=1\textwidth]{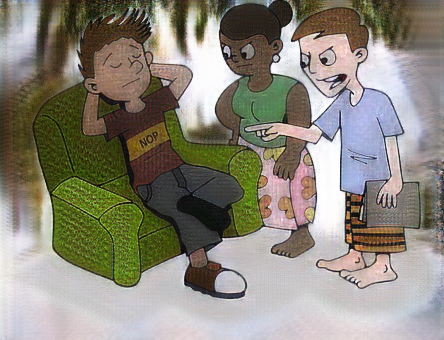}}
  \rightline{\includegraphics[width=1\textwidth]{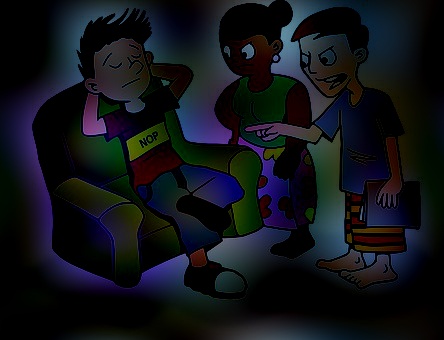}}
  \rightline{\includegraphics[width=1\textwidth]{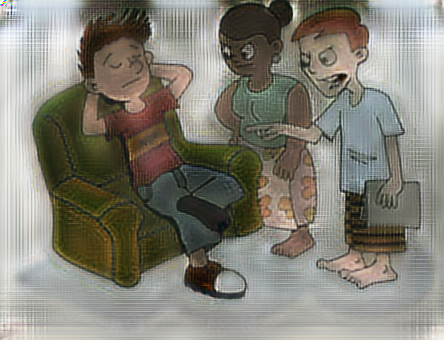}}
  \centerline{\footnotesize{(d)}}\medskip
\end{minipage}
\caption{Example images of style transfer from Clipart1k (source domain) to PascalVOC (target domain) using different generation methods, including CycleGAN (second row), FDA (third row) and our trained CUT (last row).}
\label{figCompare}
\end{figure}

Specifically, we here adopt a more superior unpaired image translator CUT~\cite{park2020contrastive} for faster and more robust scene transfer. To illustrate its advantages over other scene transfer methods (e.g., CycleGAN or FDA), we have added the comparison of transferred images to reflect the significant perceptual differences. Some examples are shown in Fig.~\ref{figCompare}. We represent the translated target-like images from source images $\mathbf{I}^s$ as $\mathbf{I}^s_{f}$, and source-like images from target images $\mathbf{I}^t$ as $\mathbf{I}^t_{f}$, respectively. Please note that image pairs $(\mathbf{I}^s, \mathbf{I}^s_{f})$ and $(\mathbf{I}^t, \mathbf{I}^t_{f})$ always appear concurrently during training. 

\subsection{Remedying Cross-Domain Discrepancy}

After generating $\mathbf{I}^s_{f}$ and $\mathbf{I}^t_{f}$, to remedy the student model with cross-domain discrepancy, we add a new supervised branch with target-like images $\mathbf{I}^s_{f}$ as the input (refer the yellow flow in Fig.~\ref{fig1}), and train them exactly the same as source images $\mathbf{I}^s$ (refer the red flow in Fig.~\ref{fig1}). Similar to Eqn.~\ref{eqn1}, the corresponding loss function is defined as follows:
\begin{equation}
  \mathcal{L}^{\dag}_{det}(\mathbf{I}^s_f, \mathcal{B}^s, \mathcal{C}^s) = \mathcal{L}_{box}(\mathcal{B}^s; \mathbf{I}^s_f) + \mathcal{L}_{cls, obj}(\mathcal{C}^s; \mathbf{I}^s_f) ~
  \label{eqn4}
\end{equation}

For the teacher model, to make it learn the global image-level features of the source domain, we replace the original input target images $\mathbf{\hat{I}}^t$ with source-like fake images $\mathbf{\hat{I}}^t_{f}$ (refer the blue flow in Fig.~\ref{fig1}). The unlabeled target images $\mathbf{\bar{I}}^t$ for training the student model keep unchanging (refer the green flow in Fig.~\ref{fig1}). Thus, we update the distillation loss in Eqn.~\ref{eqn3} into:
\begin{equation}
  \mathcal{L}^{\dag}_{dis}(\mathbf{\hat{I}}^t_f, \mathbf{\bar{I}}^t) = \mathcal{L}_{det}(\mathbf{\bar{I}}^t, \mathcal{G}_\mathcal{B}[\mathcal{F}_\mathcal{B}(\mathbf{\hat{I}}^t_f)], \mathcal{G}_\mathcal{C}[\mathcal{F}_\mathcal{C}(\mathbf{\hat{I}}^t_f)]) ~
  \label{eqn5}
\end{equation}

The relationship between Eqn.~\ref{eqn4} and~\ref{eqn5} is established by EMA parameter update in MT model. In this way, the learned teacher model will not dramatically tend to only be good at predicting objects in target domains. Also, the training of student model will gradually approach the real target domain with weak supervision of pseudo-labels from filtered predictions of $\mathbf{\hat{I}}^t_{f}$. Although not that accurate, these pseudo-labels play an alternative role for promoting fine-grained instance-level adaptation.

\subsection{Consistency Loss Function}

Although the source and target-like paired images $(\mathbf{I}^t, \mathbf{I}^t_{f})$ fed into the student model have different scene-level data distributions, they belong to the same label space. Ideally, a rational hypothesis is that the outputs of the student model feeding with both domain images should be consistent. Therefore, to ensure that their outputs are as close as possible, we can add a new constraint to the corresponding two branches. Intuitively, we have three choices: 1) apply the intermediate supervision between corresponding feature maps; 2) apply the error constraint between final predictions; 3) combine above both strategies.

The intermediate supervision strategy is initially proposed and studied by convolutional pose machines (CPM)~\cite{wei2016convolutional} for the single person pose estimation task. It is proven reasonable in addressing vanishing gradients in supervised training. However, we here need an UDA penalization. In other words, we do not want the intermediate features of the student model to be similar after the input of $\mathbf{I}^t$ and $\mathbf{I}^t_{f}$, but only expect the predictions to be as consistent as possible. The intermediate supervision here may be an excessive constraint. Thus, we select the second constraint by calculating the L2 distance between two final outputs. We call it consistency loss produced as follows:
\begin{equation}
  \mathcal{L}_{con} = \| \mathcal{L}_{det}(\mathbf{I}^s, \mathcal{B}^s, \mathcal{C}^s) - \mathcal{L}^{\dag}_{det}(\mathbf{I}^s_f, \mathcal{B}^s, \mathcal{C}^s) \|_2~
  \label{eqn6}
\end{equation}
where we can also use the L1 loss. We will analyze which one is better in our ablation studies. The consistency loss seemingly can redress cross-domain biases of objectness and classification. Its effectiveness will also be proven by experiments.

\subsection{Overall Optimization}

The framework of our SSDA-YOLO is shown in Fig.~\ref{fig1}. The target-like images $\mathbf{I}^s_{f}$ and source-like images $\mathbf{I}^t_{f}$ are generated offline by corresponding pre-trained CUT models. During inference, we only need to adopt the finely trained student model and take the target image as the single input. Our model can be trained in an end-to-end manner by optimizing all related losses jointly. The overall loss function is written as:
\begin{equation}\footnotesize
  \mathcal{L} = \mathcal{L}_{det}(\mathbf{I}^s, \mathcal{B}^s, \mathcal{C}^s) + \mathcal{L}^{\dag}_{det}(\mathbf{I}^s_f, \mathcal{B}^s, \mathcal{C}^s) + \alpha \cdotp \mathcal{L}^{\dag}_{dis}(\mathbf{\hat{I}}^t_f, \mathbf{\bar{I}}^t) + \beta \cdotp \mathcal{L}_{con} ~
  \label{eqn7}
\end{equation}
where the $\alpha$ and $\beta$ are the trade-off hyper-parameters, and determined by experiments.


\section{Experiments}

\subsection{Training Configuration}

In all experiments, we choose YOLOv5 with {\bf L}arge parameters (YOLOv5-L) as the detector for its comparable detection accuracy with state-of-the-art object detection methods in about real-time. All training and testing images are padded and resized in shape $(960, 960, 3)$. During training, each batch consists of two pairs of images: $(\mathbf{I}^s, \mathbf{I}^s_{f})$ with labels and $(\mathbf{I}^t, \mathbf{I}^t_{f})$ without labels. We can set the batchsize to 10 in one 24GB GTX3090 GPU. Total epochs are 200. The $\gamma$ in EMA for the teacher model is set to $0.99$. In filters $\mathcal{G}_\mathcal{B}[\cdot]$ and $\mathcal{G}_\mathcal{C}[\cdot]$, we set the IoU threshold $\tau_{box}=0.3$ and the category score threshold $\tau_{cls}=0.8$. Other not mentioned settings keep consistent with the setup in YOLOv5 \cite{jocher2020yolov5}. For the hyper-parameters in SSDA-YOLO, we set $\alpha=0.005$ and $\beta=2.0$ in the overall loss $\mathcal{L}$. And the consistency loss $\mathcal{L}_{con}$ adopts L2 distance. These selected settings will be discussed in our ablation studies.

\subsection{Transfer Experiment Design}

Following DA-Faster~\cite{chen2018domain} and other state-of-the-art DAOD methods, we select datasets including PascalVOC~\cite{everingham2010pascal}, Clipart1k~\cite{inoue2018cross}, Cityscapes~\cite{cordts2016cityscapes}, and Foggy Cityscapes~\cite{sakaridis2018semantic} to validate the effectiveness and superiority of our approach. Among them, we implement the experiment {\bf PascalVOC$\mapsto$ Clipart1k} for comparing the real to virtual adaptation, and execute the experiment {\bf Cityscapes$\mapsto$Foggy Cityscapes} for evaluating the normal to adverse weather adaptation. Besides, to test the universality of our method in real application, we conduct the evaluation of domain adaptive behavior detection on self-made yawning datasets using various K-12 course videos. This transfer case is named {\bf Source Classroom$\mapsto$Target Classroom}. Below are details.


\begin{table*}[ht]\scriptsize 
\setlength\tabcolsep{2pt}
    \caption{The transfer results of {\bf PascalVOC$\mapsto$Clipart1k}. The bold font means the best result. The underline font means the secondly best result.}
    \centering
    \begin{tabular}{l|cccccccccccccccccccc|c|c|c}
        	\Xhline{1pt}
		{Method} & {aero} & {bcycle} & {bird} & {boat} & {bottle} & {bus} & {car} & {cat} & {chair} & {cow} & {table} & {dog} & {hrs} & {bike} & {prsn} & {plnt} & {sheep} & {sofa} & {train} & {tv} & {mAP} & {Gain} & {Rel.} \\
        	\Xhline{1.2pt}
	 	 Source Only & 26.0 & 45.9 & 23.2 & 22.1 & 20.1 & 51.7 & 29.8 & 9.4 & 34.6 & 13.6 & 30.1 & 0.9 & 33.7 & 50.0 & 37.2 & 46.2 & 18.9 & 6.7 & 34.1 & 20.5 & 27.7 & +0.0 & --- \\
		\hline \rowcolor{gray!20}
     	 	SWDA~\cite{saito2019strong} & 26.2  & 48.5 & 32.6 & 33.7 & 38.5 & 54.3 & 37.1 & 18.6 & 34.8 & 58.3 & 17.0 & 12.5 & 33.8 & 65.5 & 61.6 & 52.0 & 9.3 & 24.9 & 54.1 & 49.1 & 38.1 & +10.4 & 84.7\% \\
		\hline \rowcolor{gray!20}
		SCL~\cite{shen2019scl} & {\bf 44.7} & 50.0 & 33.6 & 27.4 & 42.2 & 55.6 & 38.3 & \underline{19.2} & 37.9 & 69.0 & 30.1 & 26.3 & 34.4 & 67.3 & 61.0 & 47.9 & 21.4 & 26.3 & 50.1 & 47.3 & 41.5 & +13.8 & 92.2\% \\
		\hline \rowcolor{gray!20}
		DM~\cite{kim2019diversify} & 25.8 & 63.2 & 24.5 & 42.4 & 47.9 & 43.1 & 37.5 & 9.1 & 47.0 & 46.7 & 26.8 & 24.9 & 48.1 & 78.7 & 63.0 & 45.0 & 21.3 & 36.1 & 52.3 & 53.4 & 41.8 & +14.1 & 92.9\% \\
		\hline \rowcolor{gray!20}
		CRDA~\cite{xu2020exploring} & 28.7 & 55.3 & 31.8 & 26.0 & 40.1 & 63.6 & 36.6 & 9.4 & 38.7 & 49.3 & 17.6 & 14.1 & 33.3 & 74.3 & 61.3 & 46.3 & 22.3 & 24.3 & 49.1 & 44.3 & 38.3 & +10.6 & 85.1\% \\
		\hline \rowcolor{gray!20}
		HTCN~\cite{chen2020harmonizing} & 33.6 & 58.9 & 34.0 & 23.4 & 45.6 & 57.0 & 39.8 & 12.0 & 39.7 & 51.3 & 21.1 & 20.1 & 39.1 & 72.8 & 63.0 & 43.1 & 19.3 & 30.1 & 50.2 & 51.8 & 40.3 & +12.6 & 89.6\% \\
		\hline \rowcolor{gray!20}
		MEAA~\cite{nguyen2020domain} & 31.3 & 53.5 &  {\bf 38.0} & 17.8 & 38.5 & \underline{69.9} & 38.2 & {\bf 23.8} & 38.3 & 58.1 & 14.6 & 18.1 & 33.8 & {\bf 88.1} & 60.3 & 42.1 & 7.8 & 30.8 & {\bf 61.1} & {\bf 58.7} & 41.1 & +13.4 & 91.3\% \\
		\hline \rowcolor{gray!20}
		ATF~\cite{he2020domain} & 41.9 & 67.0 & 27.4 & 36.4 & 41.0 & 48.5 & 42.0 & 13.1 & 39.2 & \underline{75.1} & 33.4 & 7.9 & 41.2 & 56.2 & 61.4 & 50.6 & \underline{42.0} & 25.0 & 53.1 & 39.1 & 42.1 & +14.4 & 93.6\% \\
		\hline \rowcolor{gray!20}
		I$^3$Net~\cite{chen2021i3net} & 30.0 & 67.0 & 32.5 & 21.8 & 29.2 & 62.5 & 41.3 & 11.6 & 37.1 & 39.4 & 27.4 & 19.3 & 25.0 & 67.4 & 55.2 & 42.9 & 19.5 & 36.2 & 50.7 & 39.3 & 37.8 & +10.1 & 84.0\% \\
		\hline \rowcolor{gray!20}
		PF-ATF~\cite{he2021partial} & 35.6 & 59.9 & 31.6 & 32.7 & 44.1 & 49.4 & 36.8 & 18.4 & 40.3 & {\bf 79.3} & \underline{37.5} & 16.6 & {\bf 48.5} & 59.9 & 60.2 & 50.3 & 35.6 & 23.2 & 49.4 & 46.9 & 42.8 & +15.1 & 95.1\% \\
		\hline \rowcolor{gray!20}
		UMT~\cite{deng2021unbiased} & 39.6 & 59.1 & 32.4 & 35.0 & 45.1 & 61.9 & 48.4 & 7.5 & 46.0 & 67.6 & 21.4 & \underline{29.5} & \underline{48.2} & 75.9 & \underline{70.5} & 56.7 & 25.9 & 28.9 & 39.4 & 43.6 & 44.1 & +16.4 & \underline{98.0\%} \\
		\hline \rowcolor{gray!20}
		TIA~\cite{zhao2022task} & \underline{42.2} & 66.0 & \underline{36.9} & 37.3 & 43.7 & {\bf 71.8} & \underline{49.7} & 18.2 & 44.9 & 58.9 & 18.2 & 29.1 & 40.7 & \underline{87.8} & 67.4 & 49.7 & 27.4 & 27.8 & \underline{57.1} & 50.6 & {\bf 46.3} & {\bf +18.6} & {\bf 102.9\%} \\
		\hline
		Oracle & 33.3 & 47.6 & 43.1 & 38.0 & 24.5 & 82.0 & 57.4 & 22.9 & 48.4 & 49.2 & 37.9 & 46.4 & 41.1 & 54.0 & 73.7 & 39.5 & 36.7&  19.1 & 53.2 & 52.9 & 45.0 & +17.3 & 100.0\% \\
		\hline \rowcolor{gray!20}
		$Base_{DC}$(ours, Faster R-CNN) & 25.4 & 41.2 & 32.2 & 39.1 & 32.6 & 63.4 & {\bf 57.7} & 16.9 & 47.5 & 46.6 & 46.0 & {\bf 49.3} & 38.7 & 57.5 & {\bf 70.6} & 42.1 & {\bf 47.3} & 31.5 & 40.5 & 46.9 & 43.6 & +15.9 & 96.9\% \\
		\Xhline{1.2pt}
		Source Only* & 16.5 & 54.5 & 21.2 & 25.8 & 36.3 & 38.7 & 20.8 & 8.8 & 53.7  & 6.3 & 22.0 & 5.7 & 28.4 & 31.0 & 36.1 & 51.5 & 17.0 & 16.9 & 19.9 & 39.0 & 27.5 & +0.0 & --- \\
		\hline \rowcolor{gray!20}
		$Base$(ours, YOLOv5-L) & 25.3 & 66.8 & 26.5 & {\bf 50.1} & {\bf 62.1} & 47.5 & 39.3 & 6.7 & 57.6 & 46.4 & 33.3 & 15.2 & 23.4 & 39.9 & 57.2 & 57.8 & 20.5 & {\bf 45.0} & 37.0 & 50.5 & 40.4 & +12.9 & 88.2\% \\
		\hline \rowcolor{gray!20}
		$Base_{D}$(ours, YOLOv5-L) & 32.2 & 66.1 & 29.5 & 47.7 & 58.0 & 43.6 & 36.5 & 9.5 & \underline{63.8} & 47.1 & 35.8 & 12.4 & 30.3 & 44.9 & 58.4 & \underline{59.2} & 24.0 & 37.0 & 39.9 & 52.5 & 41.2 & +13.7 & 90.0\% \\
		\hline \rowcolor{gray!20}
		$Base_{C}$(ours, YOLOv5-L) & 33.0 & \underline{72.9} & 30.9 & \underline{49.6} & 60.3 & 35.4 & 40.2 & 12.4 & 60.7 & 52.6 & 33.8 & 17.7 & 35.7 & 45.4 & 58.3 & {\bf 60.5} & 22.8 & 31.6 & 34.6 & 55.3 & 42.2 & +14.7 & 92.1\% \\
		\hline \rowcolor{gray!20}
		$Base_{DC}$(ours, YOLOv5-L) & 30.2 & {\bf 76.6} & 33.2 & 48.4 & \underline{60.7} & 41.8 & 39.7 & 8.9 & {\bf 64.8} & 51.0 & {\bf 38.7} & 25.5 & 36.8 & 44.5 & 59.9 & 57.3 & 18.4 & \underline{41.4} & 49.2 & \underline{58.6} & \underline{44.3} & \underline{+16.8} & 96.7\% \\
		\hline
		Oracle* & 21.7 & 74.3 & 34.6 & 39.8 & 47.1 & 74.8 & 57.8 & 22.3 & 47.4 & 52.7 & 34.1 & 39.1 & 29.8 & 58.7 & 75.3 & 46.2 & 45.4 & 27.0 & 48.0 & 41.0 & 45.8 & +18.3 & 100.0\% \\
		\Xhline{1.2pt}
    \end{tabular}
    \label{tab1}
\end{table*}

\subsubsection{Real to Virtual Adaptation}

{\bf Datasets:} Following~\cite{inoue2018cross,saito2019strong,chen2021i3net,deng2021unbiased}, we combine the PascalVOC 2007 and 2012 datasets (totally 16,551 images) as the source domain, and take the Clipart1k dataset as the target domain. The Clipart1k contains $1,000$ images from the same 20 classes as the PascalVOC dataset, and is split equally into the training and test set. We use the labeled source training set and unlabeled target training images for adaptive training. The target test set is held out for validation. The same setup is true for the following experiments.

{\bf Results:} We choose 11 representative methods as in Table~\ref{tab1} to compare. They are all based on the Faster R-CNN. Thus, we re-executed the Source Only\footnote{The Source Only indicates training with labeled source images and directly testing on the target data without domain adaptation.} and Oracle\footnote{The Oracle indicates training and testing with labeled target images.} experiments using the conventional YOLOv5 (with * marker). Following them, we report the average precision (AP) at IoU=0.5 of each class as well as the mean AP over all classes in Table~\ref{tab1} for object detection on the Clipart1k test set. Additionally, both the mAP gain (Gain) and the relative UDA improvement (Rel.) wrt. the oracle mAP are provided for comparing among different basic detectors (e.g., Faster R-CNN and YOLOv5).

Particularly, we get a mAP of 40.4 with the basic SSDA-YOLO framework $Base$ by only using cross-generated fake images, yet without adding the distillation loss for remedying the discrepancy ($\alpha=0$) or adding the consistency loss ($\beta=0$). Then, by using the distillation loss, $Base_{D}$ improves the mAP into 41.2. Or only adding the consistency loss, $Base_{C}$ could obtain a mAP of 42.2. Finally, our full model $Base_{DC}$ reaches 44.3 mAP and 16.8 Gain (secondly best) on the Clipart1k test set, which is comparable with the state-of-the-art performances 44.1 in UMT~\cite{deng2021unbiased} and 46.3 in TIA~\cite{zhao2022task}. However, our Faster R-CNN based method did not perform as well as YOLOv5 based one comparing to UMT and TIA. This may somewhat disclose the importance of utilizing a more robust and advanced basic detector for tackling DAOD tasks. Besides, neither of our two methods based on YOLOv5 or Faster R-CNN obtained better Rel. than UMT and TIA. This may reflect our domain adaptive modules still have gaps to be improved. However, we should also focus on the achieved comparable absolute mAP result and advantage of super faster inference of YOLOv5 which is friendly to real applications.



\begin{table}[]\tiny  
\setlength\tabcolsep{1.8pt}
    \caption{The transfer results of {\bf Cityscapes$\mapsto$Foggy Cityscapes}. The bold font means the best result.}
    \centering
    \begin{tabular}{l|c|cccccccc|c|c|c}
	 \Xhline{1.2pt}
 	{Method} & {Detector} & {bus} & {bicycle} & {car} & {mcycle} & {person} & {rider} & {train} & {truck} & {mAP} & Gain & Rel. \\
	 \Xhline{1.2pt}
	 Source Only & Faster R-CNN & 24.7 & 29.0 & 27.2 & 16.4 & 24.3 & 31.5 & 9.1 & 12.1 & 21.8 & +0.0 & --- \\
	 \hline \rowcolor{gray!20}
	 DA-Faster~\cite{chen2018domain} & Faster R-CNN & 35.3 & 27.1 & 40.5 & 20.0 & 25.0 & 31.0 & 20.2 & 22.1 & 27.6 & +5.6 & 60.5\% \\
	 \hline \rowcolor{gray!20}
	 MAF~\cite{he2019multi} & Faster R-CNN & 39.9 & 33.9 & 43.9 & 29.2 & 28.2 & 39.5 & 33.3 & 23.8 & 34.0 & +12.2 & 74.6\% \\
	 \hline \rowcolor{gray!20}
	 SWDA~\cite{saito2019strong} & Faster R-CNN & 36.2 & 35.3 & 43.5 & 30.0 & 29.9 & 42.3 & 32.6 & 24.5 & 34.3 & +12.5 & 75.2\% \\
	 \hline \rowcolor{gray!20}
 	 DM~\cite{kim2019diversify} & Faster R-CNN & 38.4 & 32.2 & 44.3 & 28.4 & 30.8 & 40.5 & 34.5 & 27.2 & 34.6 & +12.8 & 75.9\% \\
	 \hline \rowcolor{gray!20}
	 NLDA~\cite{khodabandeh2019robust} & Faster R-CNN & 35.1 & 42.2 & 49.2 & 30.2 & 45.3 & 27.0 & 26.9 & 36.0 & 36.5 & +14.7 & 80.0\% \\
	 \hline \rowcolor{gray!20}
	 SCL~\cite{shen2019scl} & Faster R-CNN & 41.8 & 36.2 & 44.8 & 33.6 & 31.6 & 44.0 & 40.7 & 30.4 & 37.9 & +16.1 & 83.1\% \\
	 \hline \rowcolor{gray!20}
 	 CRDA~\cite{xu2020exploring} & Faster R-CNN & 45.1 & 34.6 & 49.2 & 30.3 & 32.9 & 43.8 & 36.4 & 27.2 & 37.4 & +15.6 & 82.0\% \\
	 \hline \rowcolor{gray!20}
	 ATF~\cite{he2020domain} & Faster R-CNN & 43.3 & 38.8 & 50.0 & 33.4 & 34.6 & 47.0 & 38.7 & 23.7 & 38.7 & +16.9 & 84.7\% \\
	 \hline \rowcolor{gray!20}
 	 HTCN~\cite{chen2020harmonizing} & Faster R-CNN & 47.4 & 37.1 & 47.9 & 32.3 & 33.2 & 47.5 & 40.9 & 31.6 & 39.8 & +18.0 & 87.3\% \\
	 \hline \rowcolor{gray!20}
	 EPMDA~\cite{hsu2020every} & FCOS & 44.9 & 36.1 & 57.1 & 29.0 & 41.5 & 43.6 & 39.7 & 29.4 & 40.2 & +18.4 & 88.2\% \\
	 \hline \rowcolor{gray!20}
	 MEAA~\cite{nguyen2020domain} & Faster R-CNN & 42.7 & 36.2 & 52.4 & 33.2 & 34.2 & 48.9 & 46.0 & 30.3 & 40.5 & +18.7 & 88.8\% \\
	 \hline \rowcolor{gray!20}
	 UMT~\cite{deng2021unbiased} & Faster R-CNN & 56.5 & 37.3 & 48.6 & 30.4 & 33.0 & 46.7 & 46.8 & 34.1 & 41.7 & +19.9 & 91.4\% \\
	 \hline \rowcolor{gray!20}
	 PF-ATF~\cite{he2021partial} & Faster R-CNN & 46.6 & 39.5 & 52.8 & 33.6 & 37.9 & 49.6 & 48.7 & 27.0 & 42.0 & +20.2 & 92.1\% \\
	 \hline \rowcolor{gray!20}
	 TIA~\cite{zhao2022task} & Faster R-CNN & 52.1 & 38.1 & 49.7 & 37.7 & 34.8 & 46.3 & 48.6 & 31.1 & 42.3 & +20.5 & 92.8\% \\
	 \hline \rowcolor{gray!20}
	 MGADA~\cite{zhou2022multi} & FCOS & 53.2 & 36.9 & 61.5 & 27.9 & 43.1 & 47.3 & 50.3 & 30.2 & 43.8 & +22.0 & 96.1\% \\
	 \hline \rowcolor{gray!20}
	 SIGMA~\cite{li2022sigma} & FCOS & 50.4 & 40.6 & 60.3 & 31.7 & 44.0 & 43.9 & {\bf 51.5} & 31.6 & 44.2 & +22.6 & 96.9\% \\
	 \hline \rowcolor{gray!20}
	 PT~\cite{chen2022learning} & Faster R-CNN & 56.6 & 48.7 & 63.4 & 41.3 & 43.2 & 52.4 & 37.8 & 33.4 & 47.1 & +25.3 & 103.3\% \\
	 \hline \rowcolor{gray!20}
	 TDD~\cite{he2022cross} & Faster R-CNN & 53.0 & 49.1 & 68.2 & 38.9 & 50.7 & 53.7 & 45.1 & 35.1 & 49.2 & {\bf +27.4} & {\bf 107.9\%} \\
	 \hline
      Oracle & Faster R-CNN & 49.9 & 45.8 & 65.2 & 39.6 & 46.5 & 51.3 & 34.2 & 32.6 & 45.6 & +23.8 & 100.0\% \\
	 \hline \rowcolor{gray!20}
	 $Base_{DC}$(ours) & Faster R-CNN & 48.0 & 44.0 & 61.7 & 35.8 & 49.3 & 50.4 & 36.4 & 28.8 & 44.3 & +22.5 & 97.1\% \\
	 \Xhline{1.2pt}
      	 Source Only* & YOLOv5-L & 37.2 & 39.2 & 51.9 & 30.3 & 46.5 & 49.0 & 8.5 & 24.2 & 35.9 & +0.0 & --- \\
	 \hline \rowcolor{gray!20}
      	 $Base$ & YOLOv5-L & 55.8 & 50.8 & 71.4 & 40.9 & 57.7 & 57.9 & 46.3 & 31.3 & 51.5 & +15.6 & 90.0\% \\
	 \hline \rowcolor{gray!20}
      	 $Base_{D}$ & YOLOv5-L & 55.7 & 53.1 & 71.7 & 44.2 & 59.4 & 61.2 & 46.9 & 36.0 & 53.5 & +17.6 & 93.5\% \\
	 \hline \rowcolor{gray!20}
      	 $Base_{C}$ & YOLOv5-L & 61.9 & 52.8 & 73.6 & 47.2 & 60.2 & 59.0 & 42.9 & 37.1 & 54.3 & +18.4 & 94.9\% \\
	 \hline \rowcolor{gray!20}
      	 $Base_{DC}$ & YOLOv5-L & {\bf 63.0} & {\bf 53.6} & {\bf 74.3} & {\bf 47.4} & {\bf 60.6} & {\bf 62.1} & 48.0 & {\bf 37.8} & {\bf 55.9} & +20.0 & 97.7\% \\
	 \hline
	 	Oracle* & YOLOv5-L & 61.2 & 52.1 & 77.2 & 47.8 & 62.8 & 60.3 & 50.3 & 46.2 & 57.2 & +21.3 & 100.0\% \\ 
	 \Xhline{1.2pt}
    \end{tabular}
    \label{tab2}
\end{table}

\begin{figure}[ht]
\begin{minipage}[b]{1\linewidth}
  \centering
  \leftline{\includegraphics[width=1\textwidth]{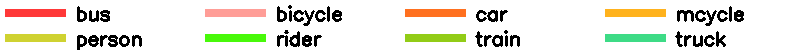}}
\end{minipage}
\begin{minipage}[b]{.325\linewidth}
  \centering
  \leftline{\includegraphics[width=1\textwidth]{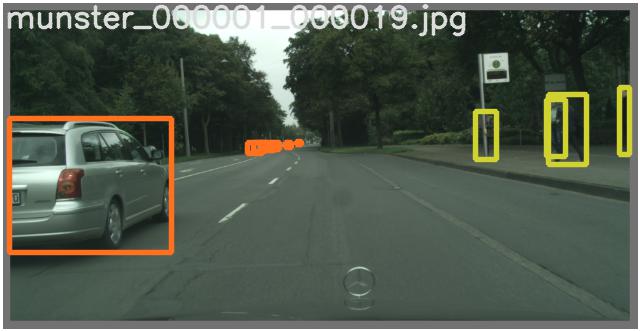}}
  \leftline{\includegraphics[width=1\textwidth]{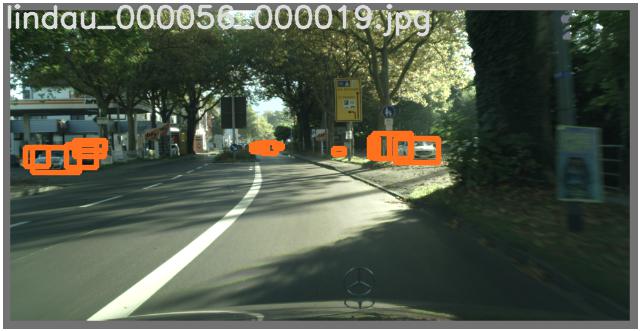}}
  \leftline{\includegraphics[width=1\textwidth]{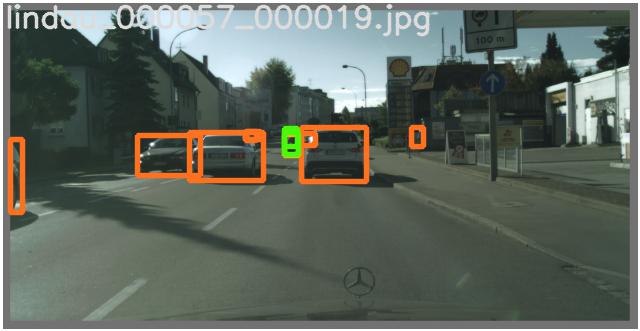}}
  \leftline{\includegraphics[width=1\textwidth]{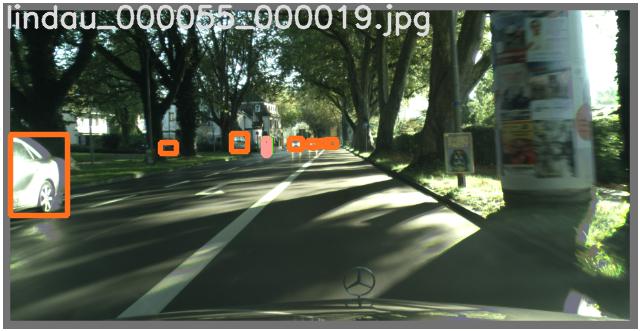}}
  \centerline{\footnotesize{(a) Oracle YOLOv5}}\medskip
\end{minipage}
\begin{minipage}[b]{.325\linewidth}
  \centerline{\includegraphics[width=1\textwidth]{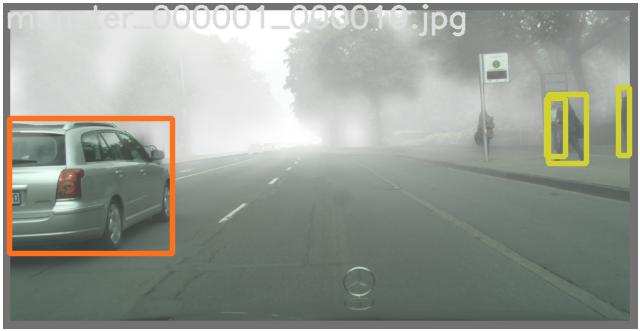}}
  \centerline{\includegraphics[width=1\textwidth]{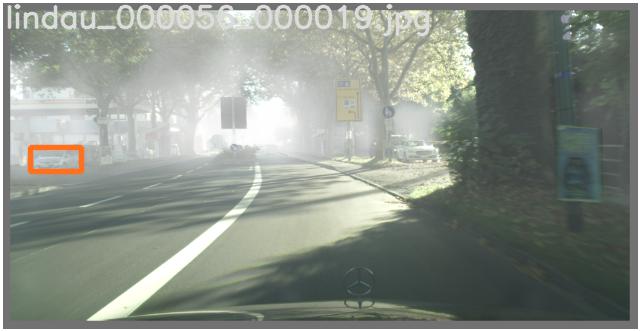}}
  \centerline{\includegraphics[width=1\textwidth]{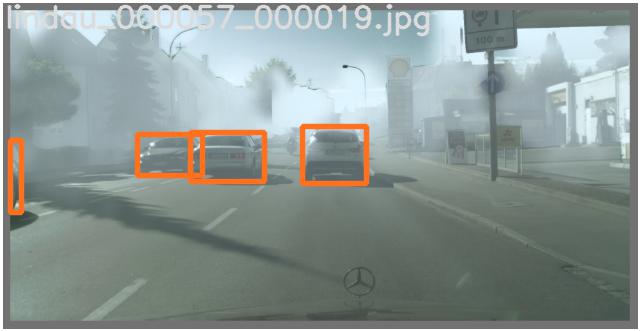}}
  \centerline{\includegraphics[width=1\textwidth]{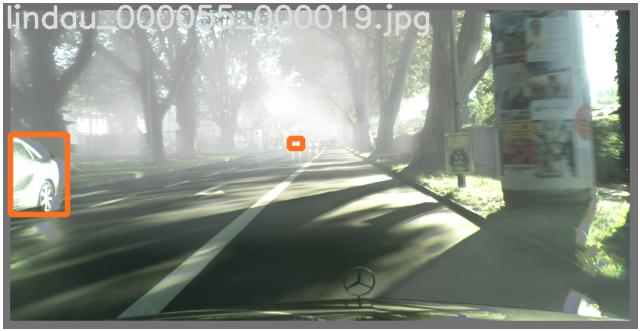}}
  \centerline{\footnotesize{(b) Source Only YOLOv5}}\medskip
\end{minipage}
\begin{minipage}[b]{.325\linewidth}
  \rightline{\includegraphics[width=1\textwidth]{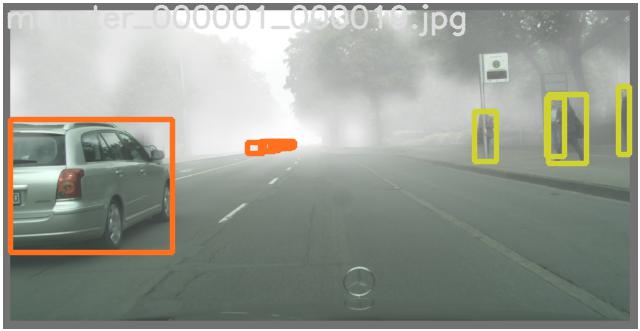}}
  \rightline{\includegraphics[width=1\textwidth]{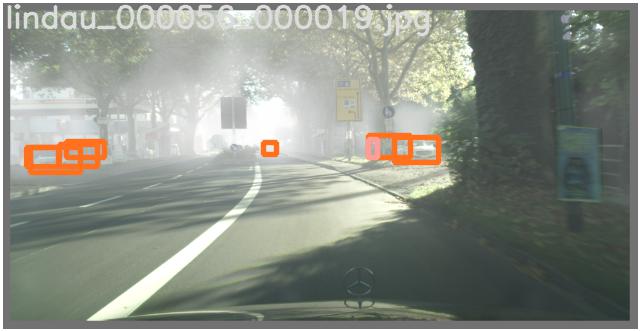}}
  \rightline{\includegraphics[width=1\textwidth]{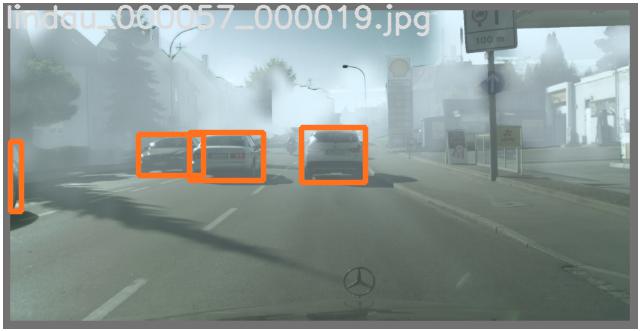}}
  \rightline{\includegraphics[width=1\textwidth]{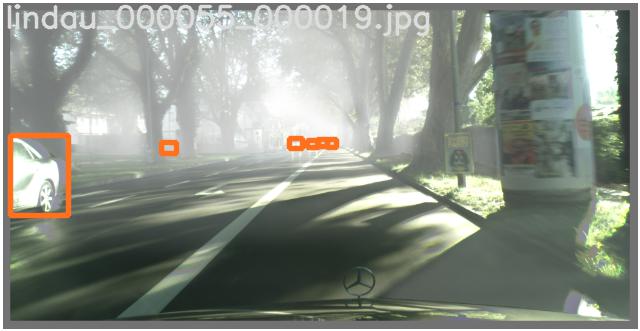}}
  \centerline{\footnotesize{(c) SSDA-YOLO}}\medskip
\end{minipage}
\caption{Visual detection examples using the original YOLOv5 model on: (a) clear images (source domain) and (b) foggy images (target domain). (c) Our proposed SSDA-YOLO applied onto foggy images.}
\label{fig2}
\end{figure}

\subsubsection{Normal to Adverse Weather Adaptation}

{\bf Datasets:} This part, we choose the Cityscapes dataset~\cite{cordts2016cityscapes} as the source domain, and the Foggy Cityscapes dataset~\cite{sakaridis2018semantic} as the target domain. The Cityscapes dataset is collected from the urban street scene captured in 50 cities. We select 8 classes for the experiment according to~\cite{chen2018domain}. The dataset contains 2,975 and 500 images in the train set and validation set, respectively. The Foggy Cityscapes is a synthetic foggy scene dataset from Cityscapes, which has exactly the same data split. We take the Cityscapes train set images with labels and the Foggy Cityscapes train set images without labels for training, and test on the validation set of Foggy Cityscapes.

{\bf Results:} We choose methods DA-Faster~\cite{chen2018domain} and 18 other representative methods to compare. Except~\cite{hsu2020every, li2022sigma, zhou2022multi} which are based on FCOS~\cite{tian2019fcos}, the others are all based on the Faster R-CNN. Again, we have retrained and got new baselines for both Source Only and Oracle using YOLOv5 (with * marker). All results of DAOD on the Foggy Cityscapes validation set are reported in Table~\ref{tab2}. Due to the efficient data augmentation of YOLOv5, the Source Only method achieves a comparable mAP value 35.9 to the recent state-of-the-art methods (e.g., EPMDA~\cite{hsu2020every} and UMT~\cite{deng2021unbiased}). Then, our basic method $Base$ is obviously better than the Source Only method. By adding the distillation loss and consistency loss, our complete model $Base_{DC}$ reaches a mAP of 55.9, which is much higher than so far best result 49.2 in TDD~\cite{he2022cross}. Some qualitative results are shown in Fig.~\ref{fig2}. For fair comparison, we should also refer the Gain and Rel. columns. The best performance of TDD did not obliterate the superiority of its adaptive strategies. Although not performing better than PT~\cite{chen2022learning} and TDD under the Rel. measure, our method either based on YOLOv5 or using Faster R-CNN outperformed methods TIA~\cite{zhao2022task}, MGADA~\cite{zhou2022multi}, and SIGMA~\cite{li2022sigma} proposed in the same year. We attribute this to the proposed adaptive strategies and the use of more advanced detectors.


\subsubsection{Cross Classroom Yawning Detection Adaptation}

\begin{figure}[]
\begin{minipage}[b]{.49\linewidth}
  \centering
  \leftline{\includegraphics[width=1\textwidth]{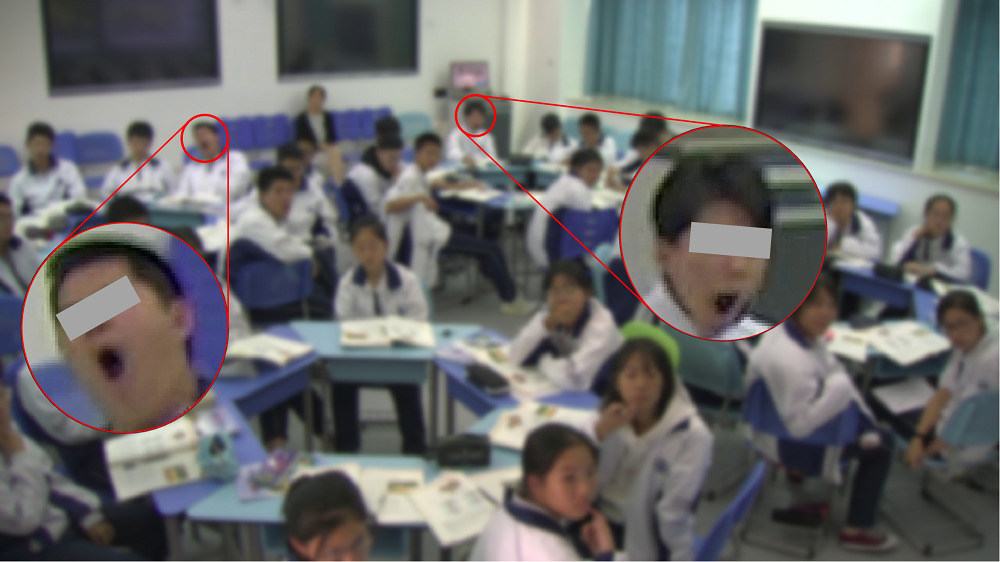}}
  \leftline{\includegraphics[width=1\textwidth]{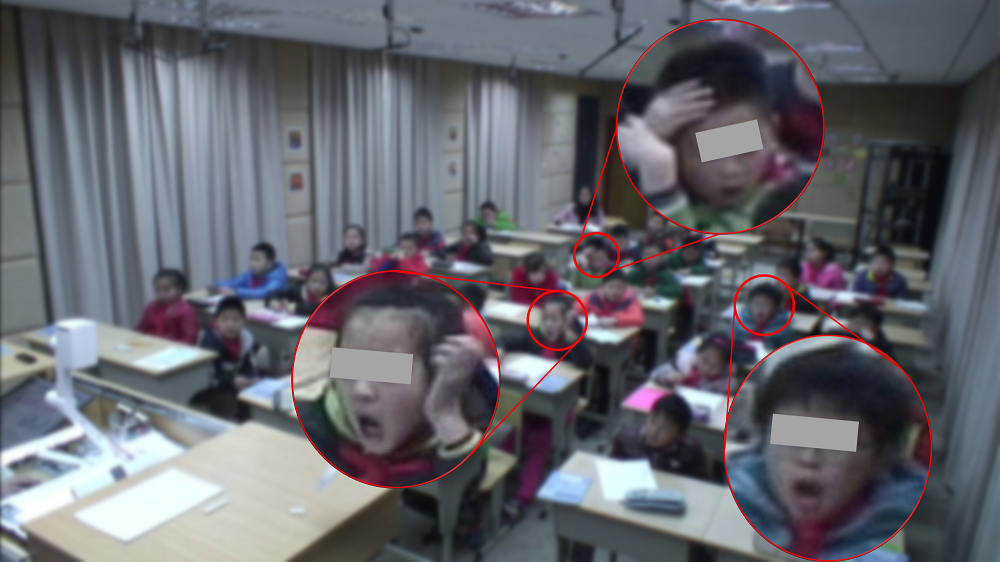}}
  \centerline{\footnotesize{(a) Source Classrooms}}\medskip
\end{minipage}
\begin{minipage}[b]{.49\linewidth}
  \rightline{\includegraphics[width=1\textwidth]{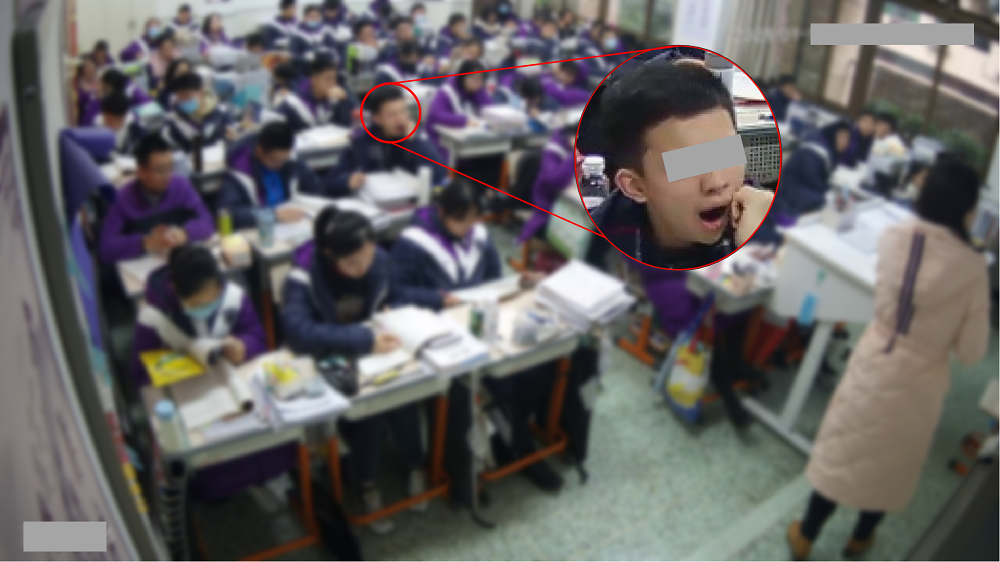}}
  \rightline{\includegraphics[width=1\textwidth]{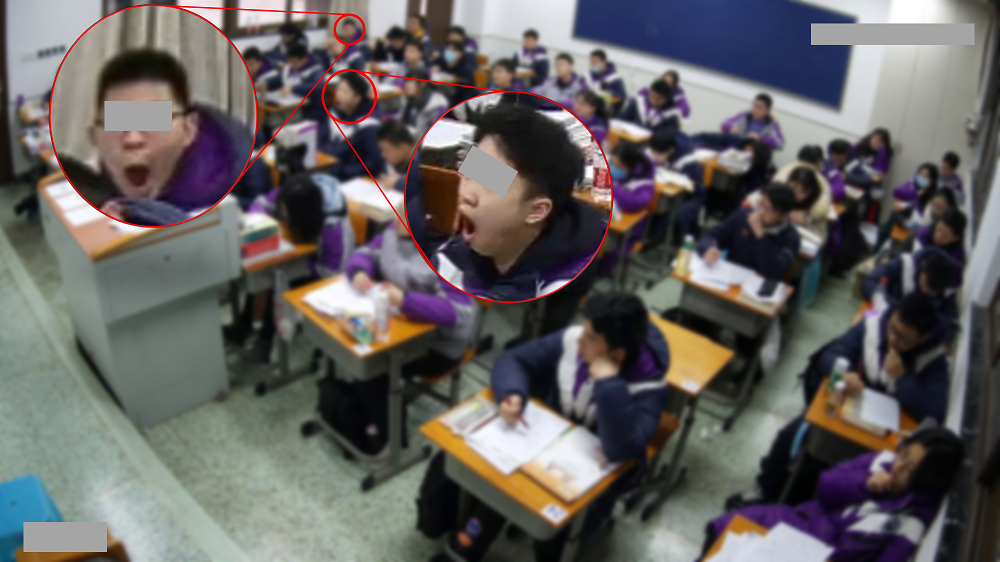}}
  \centerline{\footnotesize{(b) Target Classrooms}}\medskip
\end{minipage}
\caption{Example images with yawning behaviors are from (a) Source Classrooms and (b) Target Classrooms. Images are masked to protect privacy.}
\label{fig3}
\end{figure}

{\bf Datasets:} In our research of automatic teaching observation, we have collected about 1,300+ real course videos from 21 K-12 schools in the same city district. We define them as the source domain. Meanwhile, in another city, we have totally got 30 course videos from different classrooms in one high school. We deem these data as the target domain. Then, we sample these videos every three seconds to generate static frames and annotated the yawning behavior using bounding box. Finally, we totally obtain 12,924 valid images in the source school and 1,990 images in the target school. Two pairs of example images are shown in Fig.~\ref{fig3}. We divide the data in the two domains into train set and test set with an $8:2$ ratio, respectively.

\begin{table}[]\scriptsize  
\setlength\tabcolsep{4.5pt}
    \caption{The transfer results of {\bf Source Classroom$\mapsto$Target Classroom}.}
    \centering
    \begin{tabular}{l|ccc|c|c}
        \Xhline{1.2pt}
        {Method} & {mAP$^{0.50}$} & {mAP$^{0.75}$} & {mAP$^{.5:.95}$} & Gain & Rel. \\
        \Xhline{1.2pt}
      	  Source Only & 77.5 & 31.8 & 38.7 & +0.0 & --- \\
	 \hline\rowcolor{gray!20}
      	 $Base$(YOLOv5-L) & 77.3 & 43.6 & 43.7 & +5.0 & 67.1\% \\
	 \hline\rowcolor{gray!20}
      	 $Base_{D}$(YOLOv5-L) & 79.5 & 47.2 & 46.2 & +7.5 & 71.0\% \\
	 \hline\rowcolor{gray!20}
      	 $Base_{C}$(YOLOv5-L) & {\bf 80.4} & 50.7 & 47.6 & +8.9 & 73.1\% \\
	 \hline\rowcolor{gray!20}
      	 $Base_{DC}$(YOLOv5-L) & 79.5 & {\bf 58.8} & {\bf 52.5} & {\bf +13.8} & {\bf 80.6\%} \\
	 \hline
	 Oracle & 93.5 & 84.2 & 65.1 & +26.4 & 100\% \\ 
	 \Xhline{1.2pt}
    \end{tabular}
    \label{tab3}
\end{table}

{\bf Results:} Since we only have one category yawning, a more stringent index mAP$^{.5:.95}$ is adopted to measure the performance. Firstly, we train YOLOv5 on the train set of the source school and the target school, respectively. Then we get the results of Source Only and Oracle on the test set of the target school. Not surprisingly, as shown in Table~\ref{tab3}, even though there are a large number of yawning images in the source school, due to the domain shifts mainly caused by different classroom scenes, the final Source Only model with 38.7 mAP$^{.5:.95}$ performs rather poor compared with Oracle reaching 65.1 on the target test set. By using our methods $Base$, $Base_{D}$, $Base_{C}$ and $Base_{DC}$ in turn, the mAP$^{.5:.95}$ is improved to 43.7, 46.2, 47.6 and 52.5 respectively. Although a gap still exists comparing with the Oracle result, our proposed method can alleviate the accuracy degradation distinctly of cross-domain behavior detection in real classrooms. Besides, the relatively lower Rel. results than the former experiments illustrated the complexity of the real classroom scenario for DAOD tasks.


\subsection{Ablation Studies}

Although above transfer experiments have shown that each proposed module in SSDA-YOLO can improve the DAOD performance separately, we want to further explore the process of acquiring optimal parameters. Without loss of generality, we use YOLOv5 with {\bf S}mall parameters (YOLOv5-S) as the backbone, and {\bf PascalVOC$\mapsto$Clipart1k} as the domain transfer test case. Each training is shortened to 100 epochs for simplicity.

\begin{figure}[]
\begin{minipage}[]{0.48\linewidth}
	\centering
	\pgfplotsset{x tick label style={/pgf/number format/1000 sep=\,},
		log base 10 number format code/.code={$\pgfmathparse{10^(#1)}\pgfmathprintnumber{\pgfmathresult}$}} 
	\begin{tikzpicture}\tiny 
	\begin{semilogxaxis}[
    	xmin = 0.01, xmax = 1, ymin = 29.8, ymax = 34.2, 
    	domain=0.01:1, xtick = {0.01, 0.02, 0.05, 0.1, 0.2, 0.5, 1}, log ticks with fixed point,
	grid = both, minor tick num = 3, major grid style = {lightgray}, minor grid style = {lightgray!20},
   	width = 1.3\columnwidth, height = 1.5\columnwidth, 
   	legend style={at={(0.25,0.35)}, anchor=north west, fill=lightgray!50}
    	]
    	
	\addplot[color=black, mark = *, mark size = 2pt]
	coordinates {(0.01,30.4) (0.02,31.2) (0.05,33.6) (0.1,32.7) (0.2,32.0) (0.5,31.8) (1,30.8)};
	\node[below left] at (rel axis cs: 0.47, 0.92) {33.6};
	
	\addplot[color=black, mark=none, thick, dashed] coordinates {(0.01,32.6) (1,32.6)};
	\node[below left] at (rel axis cs: 0.98, 0.72) {\bf 32.6};
	
	\legend{\tiny{$mAP$}, \tiny{$\alpha=\beta=0$}};
	\end{semilogxaxis}
	\end{tikzpicture}
	\caption{The effect of $\alpha$ (having $\times 10$).}
	\label{alphaAS}
\end{minipage}
\begin{minipage}[]{0.48\linewidth} 
	\centering
	\pgfplotsset{x tick label style={/pgf/number format/1000 sep=\,},
		log base 10 number format code/.code={$\pgfmathparse{10^(#1)}\pgfmathprintnumber{\pgfmathresult}$}} 
	\begin{tikzpicture}\tiny 
	\begin{semilogxaxis}[
    	xmin = 0.1, xmax = 10.0, ymin = 29.8, ymax = 36.2, 
    	domain=0.1:10, xtick = {0.1, 0.2, 0.5, 1.0, 2.0, 5.0, 10.0}, log ticks with fixed point, ytick = {30,31,32,33,34,35,36},
	grid = both, minor tick num = 3, major grid style = {lightgray}, minor grid style = {lightgray!20},
   	width = 1.3\columnwidth, height = 1.5\columnwidth, 
   	legend style={at={(0.04,0.99)}, anchor=north west, fill=lightgray!50} 
    	]
    	
	\addplot[color=black, mark = square*, mark size = 2pt]
	coordinates {(0.5,30.8) (1.0,33.4) (2.0,34.7) (5.0,33.3) (10.0,31.8)};
	\node[below left] at (rel axis cs: 0.77, 0.82) {34.7};
	
	\addplot[color=black, mark = triangle*, mark size = 2.5pt]
	coordinates {(0.1,33.1) (0.2,33.2) (0.5,33.8) (1.0,31.8) (2.0,30.8)};
	\node[below left] at (rel axis cs: 0.34, 0.69) {33.8};
	
	\addplot[color=black, mark=none, thick, dashed] coordinates {(0.1,32.6) (10,32.6)};
     \node[below left] at (rel axis cs: 0.11, 0.43) {\bf 32.6};
     
	\legend{\tiny{${mAP}/L2\;Loss$}, \tiny{${mAP}/L1\;Loss$}};
	\end{semilogxaxis}
	\end{tikzpicture}
	\caption{The effect of $\beta$ and loss type.}
	\label{betaAS}
\end{minipage} 
\end{figure}

\subsubsection{Weight $\alpha$ for Distillation Loss}
To seek out the optimal parameter value of $\alpha$, we firstly set $\beta=0$ in total loss function, and sample $\alpha$ from 0.001 to 0.1 with about 2$\times$ increasing in each step. The experimental results are shown in Fig.~\ref{alphaAS}. The optimal mAP$^{.50}$ result is achieved when $\alpha=0.005$. And a smaller or larger $\alpha$ will decrease the performance of distillation loss for remedying cross-domain discrepancy. This indicates that we should not be too aggressive when adjusting the weight of distillation loss.

\subsubsection{Weight $\beta$ for Consistency Loss}
Similarly, to seek out the best value of $\beta$, we firstly set $\alpha=0$ and L2 loss in $\mathcal{L}_{con}$. We then select $\beta$ from five candidates (0.5, 1.0, 2.0, 5.0, 1.0). The weight $\beta$ is larger than $\alpha$, which is determined by the absolute value of the corresponding loss. Then, we obtained all testing results in Fig.~\ref{betaAS}. The optimal mAP is achieved when $\beta=2.0$. Moreover, to verify which kind of consistency loss is better, we repeat the above experiments except for selecting $\beta$ from (0.1, 0.2, 0.5, 1.0, 2.0) and changing $\mathcal{L}_{con}$ into L1 loss. As in Fig.~\ref{betaAS}, the highest mAP using L1 loss is 33.8 when $\beta=0.5$, which is inferior to the counterpart 34.7 of L2 loss.

\subsubsection{Validity of Consistency Loss}
\begin{figure}[]
\begin{minipage}[b]{0.48\linewidth}
  \centering
  \leftline{\includegraphics[width=1\textwidth]{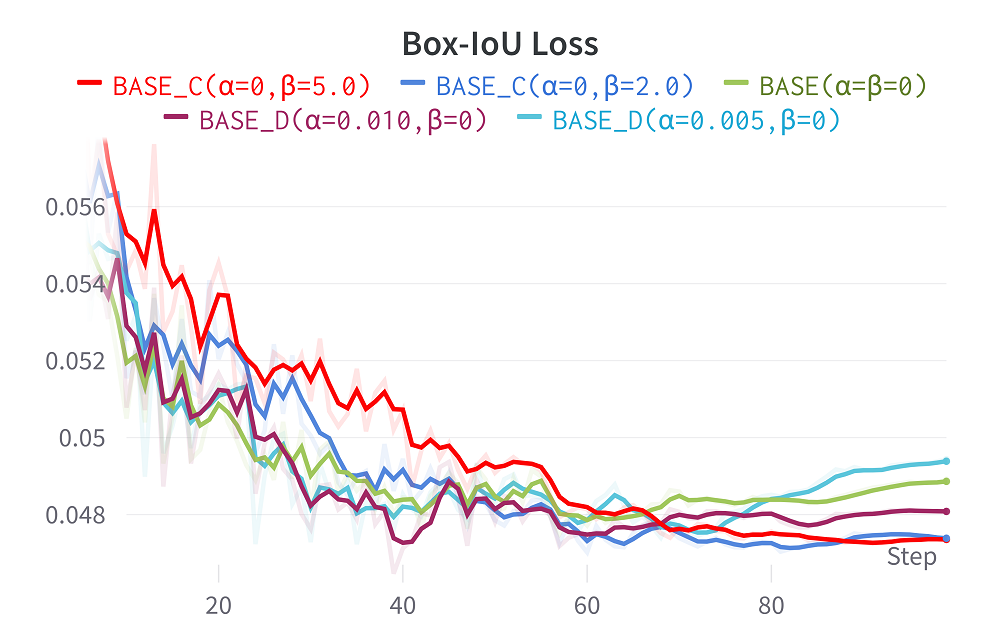}}
  \centerline{\footnotesize{(a) $\mathcal{L}_{box}$}}\medskip
\end{minipage}
\begin{minipage}[b]{0.48\linewidth}
  \rightline{\includegraphics[width=1\textwidth]{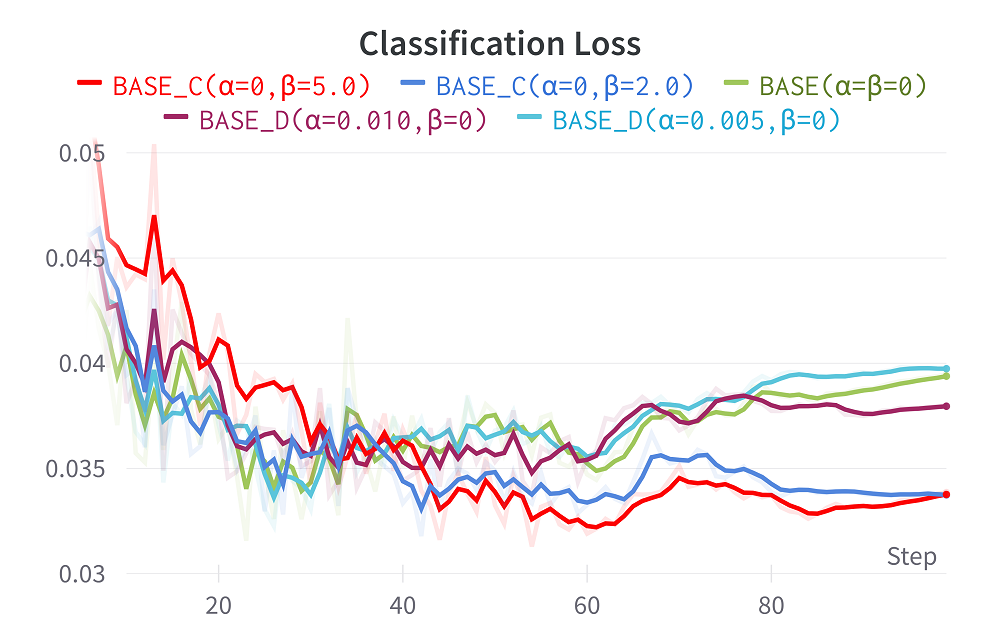}}
  \centerline{\footnotesize{(b) $\mathcal{L}_{cls}$}}\medskip
\end{minipage}
\begin{minipage}[b]{1\linewidth}
  \centerline{\includegraphics[width=0.48\textwidth]{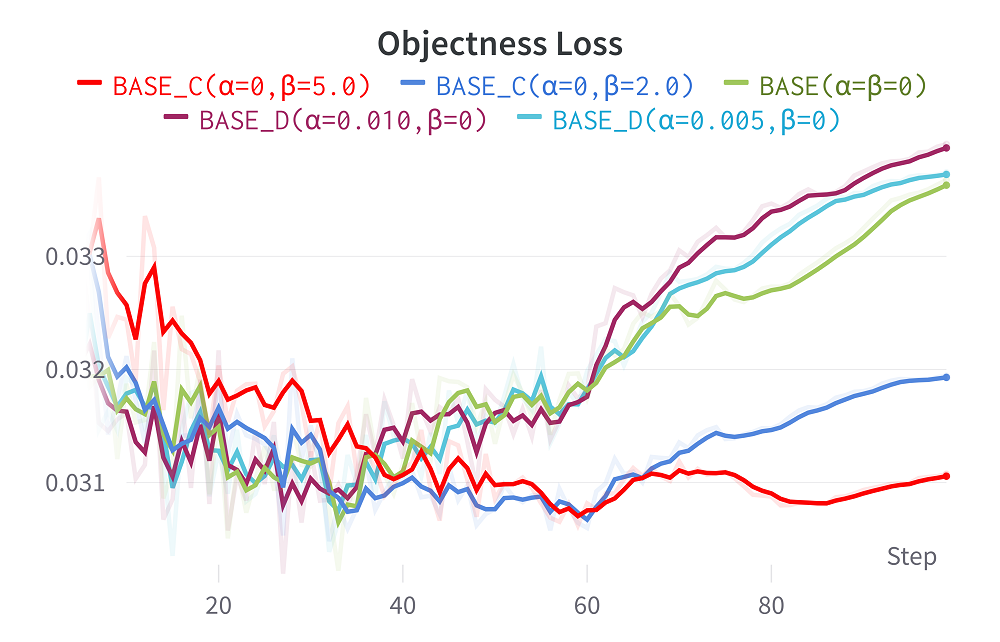}}
  \centerline{\footnotesize{(c) $\mathcal{L}_{obj}$}}\medskip
\end{minipage}
\caption{Recorded loss trends when evaluating on the target domain test set. We choose three kinds of experiment setup for comparing: 1) $Base$ without adding $\mathcal{L}_{dis}$ and $\mathcal{L}_{con}$, 2) $Base_D$ with only adding $\mathcal{L}_{dis}$, and 3) $Base_C$ with only adding $\mathcal{L}_{con}$.}
\label{conAS}
\end{figure}

Finally, we recorded the trends of $\mathcal{L}_{cls}$, $\mathcal{L}_{box}$ and $\mathcal{L}_{obj}$ on the target domain test set before and after adding the consistency loss. As we can see in Fig.~\ref{conAS}, the $\mathcal{L}_{con}$ ($Base_C$) can obviously promote the convergence of all three losses, especially the healing of objectness loss $\mathcal{L}_{obj}$. This further intuitively shows the essential effectiveness of the proposed consistency loss $\mathcal{L}_{con}$.

\section{Conclusion}

In this paper, we propose a novel semi-supervised cross-domain object detection method named SSDA-YOLO. We abandon the current less efficient outdated Faster R-CNN, which has been dominated in previous works, and introduce the more superior YOLOv5 as our backbone detector. Specifically, our framework contains three effective components. First, based on the knowledge distillation structure, we separately learn the YOLOv5 as a student network and a Mean Teacher model based teacher network to build a robust training. Next, we perform style transfer to cross-generate pseudo training images for alleviating the global domain differences. Finally, we apply a consistency loss function to correct the prediction shifts of images from different domains but with same labels. We have performed experiments on both public benchmarks and self-made yawning behavior datasets. The final results prove the effectiveness and superiority of our proposed SSDA-YOLO in real cross-domain object detection applications, and also reveal the necessity of adopting advanced detectors to bring forward DAOD researches.


\section*{Acknowledgments}
This paper is supported by NSFC (No. 62176155, 61772330), Shanghai Municipal Science and Technology Major Project (2021SHZDZX0102)

\bibliographystyle{IEEEbib}
\bibliography{mybibfile}

\end{document}